\pdfoutput=1

\documentclass[11pt]{article}
\usepackage{enumitem}
\usepackage{adjustbox}

\usepackage{subcaption} 
\usepackage{afterpage}
\PassOptionsToPackage{table}{xcolor}

\usepackage[final]{acl}
\usepackage{subcaption}

\usepackage{times}
\usepackage{latexsym}

\usepackage{amsmath}

\usepackage[T1]{fontenc}

\usepackage[utf8]{inputenc}

\usepackage{microtype}

\usepackage{inconsolata}

\usepackage{array}
\usepackage{graphicx}
\usepackage{booktabs}
\usepackage{tabularx}
\usepackage{multirow}
\usepackage[table]{xcolor}
\usepackage{multirow,makecell}
\usepackage{amsmath,amssymb}
\usepackage{fancyvrb}
\title{ThinkLess: A Training-Free Inference-Efficient Method for Reducing Reasoning Redundancy}

\author{
Gengyang Li$^{1,2}$\thanks{~~Equal contribution.} \quad 
Yifeng Gao$^{1,2}$\footnotemark[1] \quad 
Yuming Li $^{2}$ \quad
Yunfang Wu$^{1,3}$\thanks{~~Corresponding author.} \\
$^1$National Key Laboratory for Multimedia Information Processing, Peking University \\
$^2$School of Software and Microelectronics, Peking University \\
$^3$School of Computer Science, Peking University \\
\texttt{\{ligengyang, yifgao26, 2301210310\}@stu.pku.edu.cn} \quad
\texttt{wuyf}@pku.edu.cn 
}

\begin{document}
\maketitle
\begin{abstract}

While Chain-of-Thought (CoT) prompting improves reasoning in large language models (LLMs), the excessive length of reasoning tokens increases latency and KV cache memory usage, and may even truncate final answers under context limits.
We propose \textbf{ThinkLess}, an inference-efficient framework that terminates reasoning generation early and maintains output quality without modifying the model.
Atttention analysis reveals that answer tokens focus minimally on earlier reasoning steps and primarily attend to the reasoning terminator token, due to information migration under causal masking.
Building on this insight, ThinkLess inserts the terminator token at earlier positions to skip redundant reasoning while preserving the underlying knowledge transfer.
To prevent format discruption casued by early termination, ThinkLess employs a lightweight post-regulation mechanism, relying on the model's natural instruction-following ability to produce well-structured answers. Without fine-tuning or auxiliary data, ThinkLess achieves comparable accuracy to full-length CoT decoding while greatly reducing decoding time and memory consumption.

\end{abstract}

\section{Introduction}

Large language models (LLMs)~\cite{vaswani2017attention, zhang2025will} have achieved remarkable progress in natural language understanding and generation, but still struggle with tasks requiring multi-step reasoning. Chain-of-Thought (CoT) prompting~\cite{wei2022chain} has emerged as a popular approach to address this issue, enabling models to decompose problems into intermediate reasoning steps before producing an answer.

\begin{figure}[!t]
  \includegraphics[width=\linewidth]{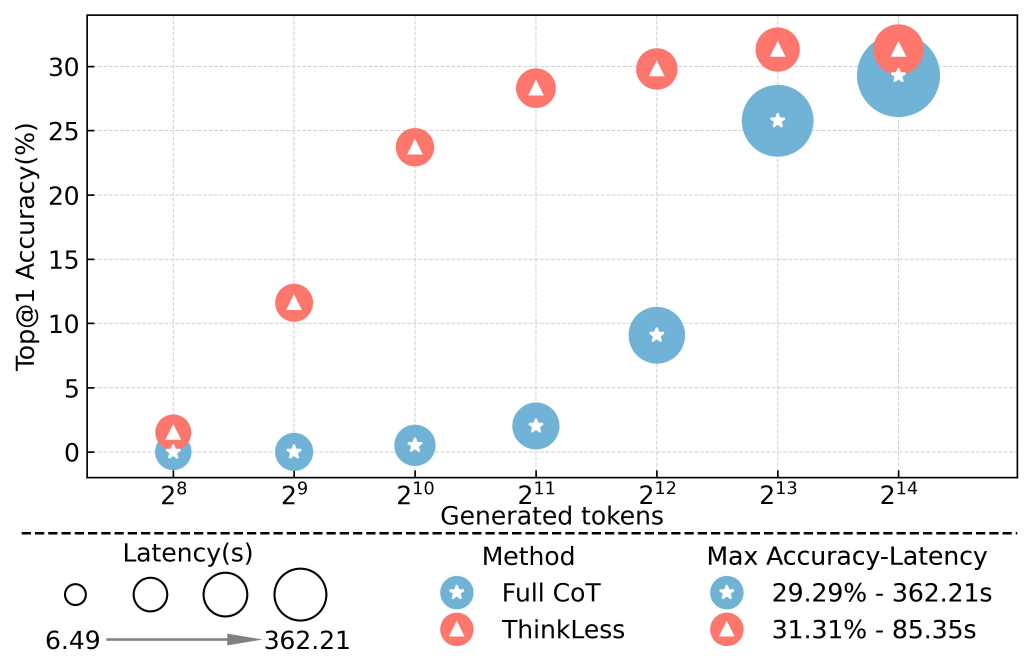}
  \caption{{\footnotesize GPQA~\cite{rein2024gpqa} accuracy of DeepSeek-R1-Distill-LLaMA-8B~\cite{guo2025deepseek} under varying token budgets. {\color{red}Red}: ThinkLess (compressed reasoning); {\color{blue}Blue}: full CoT reasoning.The left part of the legend illustrates the relationship between marker size and latency, the middle part denotes each methods, and the right part presents the maximum accuracy and corresponding latency of each method.}}
  \label{fig:illustration}
\end{figure}

While CoT improves accuracy on challenging benchmarks~\cite{zhang2022automatic, jaech2024openai}, it comes at a cost: reasoning tokens tend to be long and autoregressively generated, introducing substantial latency and memory overhead during inference. As shown in Figure\,\ref{fig:illustration}, increasing the token budget does improve accuracy--but the gains diminishes rapidly, indicating clear marginal returns. Beyond a certain point, longer generations incur significantly higher computational cost without meaningful performance improvement. In deployment scenarios where user experience and response time are critical, such overhead becomes a practical bottleneck, making blind expansion of reasoning length both inefficient and unsustainable.

Several efforts aim to improve CoT efficiency through techniques such as feedback-based refinement~\cite{yao2023react}, search and planning~\cite{bi2024forest, ye2024multi}, and iterative optimization~\cite{zhang2024self}. While effective in controlled settings, these approaches typically rely on \textit{additional training}, \textit{curated datasets}, or \textit{supervised fine-tuning (SFT)}--introducing significant engineering overhead. Moreover, their reliance on task-specific data or model customization limits generalizability, making them difficult to scale or deploy in real-world systems where flexibility, modularity, and minimal intervention are critical.

We introduce \textbf{ThinkLess}, an inference-efficient framework that reduces DeepSeek-R1~\cite{guo2025deepseek} distilled CoT reasoning overhead \textit{without any model modification or additional training}. Our key insight stems from an attention analysis: during answer generation, models rely minimally on earlier reasoning steps and focus on disproportionately on the reasoning terminator tokens (\emph{e.g.}, <$\mathrm{/think}$>). This indicates reasoning information is progressively migrated and compressed toward the end of the reasoning sequence due to causal attention~\cite{lin2025boosting}.

However, naively truncating reasoning by inserting the terminator token early often results in disrupted output formats. To solve, ThinkLess employs a lightweight output regulation that guides the model to produce well-structured responses. This is implemented simply by appending a small instruction prompt after early termination, leveraging the model’s inherent instruction-following capabilities. This post-regulation step requires no model modification or fine-tuning, yet proves essential for maintaining output consistency and restoring accuracy degraded by premature reasoning truncation.

ThinkLess achieves substantial efficiency gains. As illustrated in Figure\,\ref{fig:illustration}, ThinkLess reaches strong performance at a much lower token budget compared to full CoT decoding, and further reduces inference latency, as reflected by smaller sizes. These results demonstrate that long-form reasoning is not always necessary; with proper output regulation, shortened reasoning can retain accuracy while dramatically improving inference efficiency.

Our contributions are as follows:
\begin{itemize}
\item We present an attention-based analysis revealing that answer tokens in CoT generation attend minimally to earlier reasoning steps, indicating substantial redundancy.

\item We propose ThinkLess, a training-free early termination strategy that injects a reasoning terminator token to truncate redundant reasoning while preserving core information.

\item To mitigate format disruption caused by early termination, we introduce a lightweight \emph{output regulation} mechanism that restores structured answers using a minimal instruction prompt.

\item ThinkLess achieves comparable performance than full CoT decoding with fewer tokens and lower inference cost, offering a plug-and-play solution deployable across models and tasks.
\end{itemize}

\section{Related Work}

\subsection{LLMs Reasoning}
Reasoning is a fundamental capability of LLMs, enabling them to tackle complex multi-step tasks across diverse domains~\cite{qiao2022reasoning}. To enhance this ability, recent work has explored various prompting and architectural strategies.
Chain-of-Thought (CoT) prompting~\cite{wei2022chain} has emerged as a foundational method, guiding models to generate intermediate reasoning steps before producing final answers. This decomposition of complex problems into sub-goals significantly improves performance on arithmetic, commonsense, and scientific reasoning benchmarks~\cite{kojima2022large,feng2023towards,rein2024gpqa,lyu2023faithful}. Building on CoT, techniques such as Self-Consistency~\cite{wang2022self} aggregate multiple reasoning paths to improve robustness, while Tree-of-Thoughts~\cite{yao2023tree} introduces structured planning via tree-based exploration.
More recently, advanced frameworks like OpenAI’s o1, Alibaba’s QwQ~\cite{qwq32b}, and DeepSeek’s R1~\cite{guo2025deepseek} have extended CoT by incorporating reflective reasoning modes such as trial-and-error, backtracking, and self-correction~\cite{shinn2023reflexion}. 


\subsection{CoT Compression}
While deeper reasoning improves performance, it often comes with diminishing returns and increasing computational cost~\cite{chen2024not,wu2024comparative}. Excessively long reasoning sequences not only prolong inference but also strain memory and may even degrade output quality~\cite{liu2025understanding,liu2025there}.
Recent work has thus focused on efficient CoT generation, which falls into two broad categories: training-based compression and inference-time optimization~\cite{qu2025survey,sui2025stop}.
Training-based methods learn more compact reasoning traces through supervised fine-tuning. Some approaches compress CoT chains at the token level~\cite{han2024token,xia2025tokenskip}, dynamically adjusting reasoning length based on task difficulty~\cite{hao2024training,zhang2025lightthinker}. Others replace explicit token-level reasoning with latent or abstract representations~\cite{chen2024not,shen2025dast,qu2025optimizing}, compressing the reasoning into a hidden state or learned vector.
Inference-time methods, by contrast, improve efficiency without modifying model weights. These include Sketch-of-Thought~\cite{aytes2025sketch,xu2025chain}, which generate concise draft reasoning before producing final outputs, balancing coherence and computational cost.

Our ThinkLess, aligns with this line of inference-time CoT optimization but differs by being entirely training-free and model-agnostic, particularly for DeepSeek-R1 distilled models. Rather than compressing reasoning through learning, ThinkLess truncates redundant reasoning tokens based on attention insights and restores output quality through a lightweight post-regulation mechanism.

\section{Methodology}

We present \textbf{ThinkLess}, a \textit{training-free} framework designed to improve inference efficiency for CoT reasoning in LLMs. ThinkLess achieves this by (1) identifying redundancy in long reasoning traces via attention and hidden state analyses, and (2) introducing a lightweight termination and regulation mechanism that preserves output accuracy and format while significantly reducing decoding cost.

\subsection{CoT Bottlenecks at Inference}

\paragraph{Problem Formulation.}
Given a question $q$, LLM generates a sequence of tokens $x_{1:N}$ autoregressively, where each token $x_i$ is sampled based on the conditional probability $p(x_i \mid q, x_{<i})$. In CoT prompting, this sequence can be divided into reasoning tokens $x^{\mathrm{reason}}_{1:M}$ and answer tokens $x^{\mathrm{answer}}_{1:N}$:
\begin{equation}
\begin{split}
p(x^{\mathrm{reason}}_{1:M} \mid q) &= \prod_{i=1}^{M} p(x_i^{\mathrm{reason}} \mid q, x^{\mathrm{reason}}_{<i})
\end{split}
\end{equation}

\begin{equation}
\begin{split}
&p(x^{\mathrm{answer}}_{1:N} \mid q, x^{\mathrm{reason}}_{1:M}) =\\& \prod_{i=1}^{N} p(x_i^{\mathrm{answer}} \mid q, x^{\mathrm{reason}}_{1:M}, x^{\mathrm{answer}}_{<i}).
\end{split}
\end{equation}

\paragraph{Inference-Time Bottlenecks.}
While reasoning tokens can enhance the model’s ability to arrive at a more accurate answer during training, they introduce significant overhead during inference. Specifically, long reasoning sequences lead to increased computational costs, higher memory usage (due to the expanded KV cache~\cite{qin2025cake}), and longer response times. This is particularly problematic in applications where quick answer responses are crucial, such as interactive AI systems.

Also, long reasoning paths may consume the context budget before generating answers, rendering the reasoning benefits inaccessible. This mismatch between computation and usable output severely undermines the efficacy of CoT at inference time. We empirically observe this issue in Figure\,\ref{fig:illustration}, where the model's performance noticeably degrades when the total token length falls below $2^{13}$. One key reason is that the answer segment is often truncated due to limited context, preventing the model from fully leveraging the reasoning process it has computed.

\paragraph{Motivation.}
These challenges expose a core inefficiency in current CoT generation: even if reasoning is computed, the final answer may not be delivered due to truncation, or its benefits may be outweighed by the added inference burden.
These raises an important question: \textit{how much of the reasoning is actually needed to support answer generation?}
In Section\,\ref{subsec:analysis}, we examine the model's internal attention behavior during decoding to investigate this question more closely. Section\,\ref{subsec:solution} then presents a termination mechanism with minimal formatting disruption, enabling efficient and accurate CoT inference.

\subsection{Attention Reveals Redundancy in CoT Reasoning}
\label{subsec:analysis}

\begin{figure*}[!t]

    \centering
    
    \begin{subfigure}{0.235\textwidth}
        \centering
        \includegraphics[width=\linewidth]{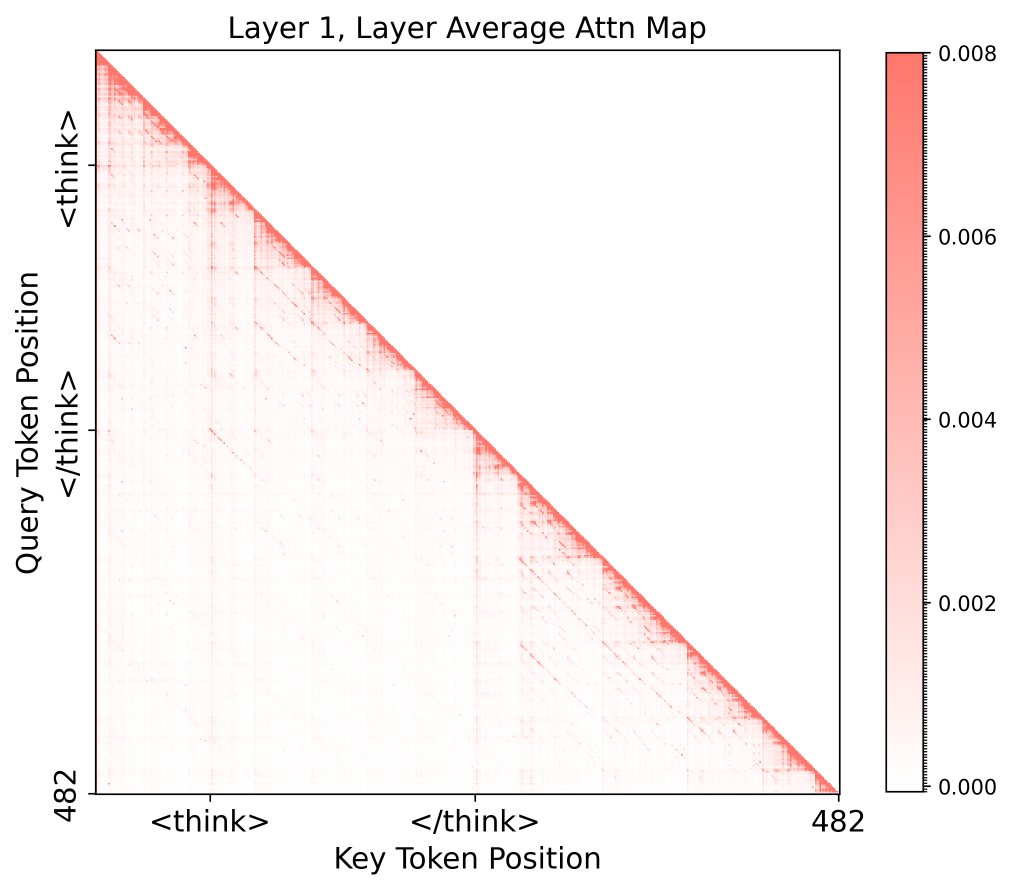}
        \caption{Layer 1}
    \end{subfigure}
    \hfill
    \begin{subfigure}{0.235\textwidth}
        \centering
        \includegraphics[width=\linewidth]{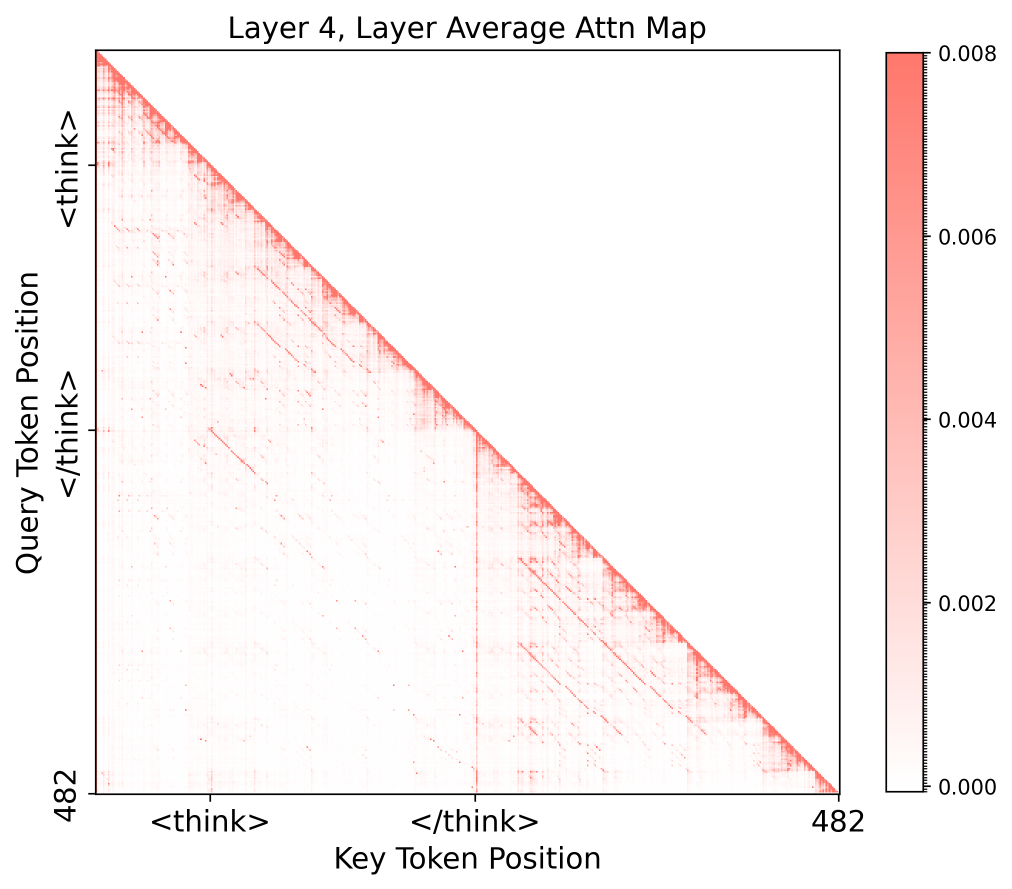}
        \caption{Layer 4}
    \end{subfigure}
    \hfill
    \begin{subfigure}{0.235\textwidth}
        \centering
        \includegraphics[width=\linewidth]{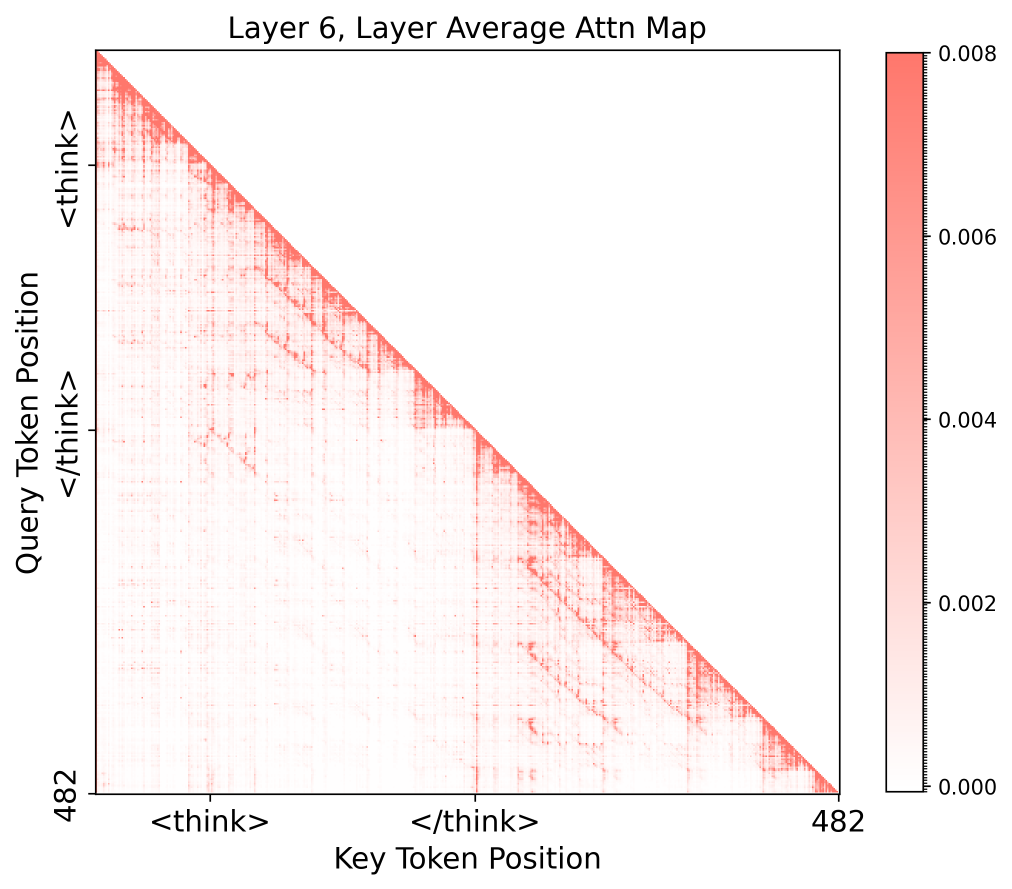}
        \caption{Layer 6}
    \end{subfigure}
    \hfill
    \begin{subfigure}{0.235\textwidth}
        \centering
        \includegraphics[width=\linewidth]{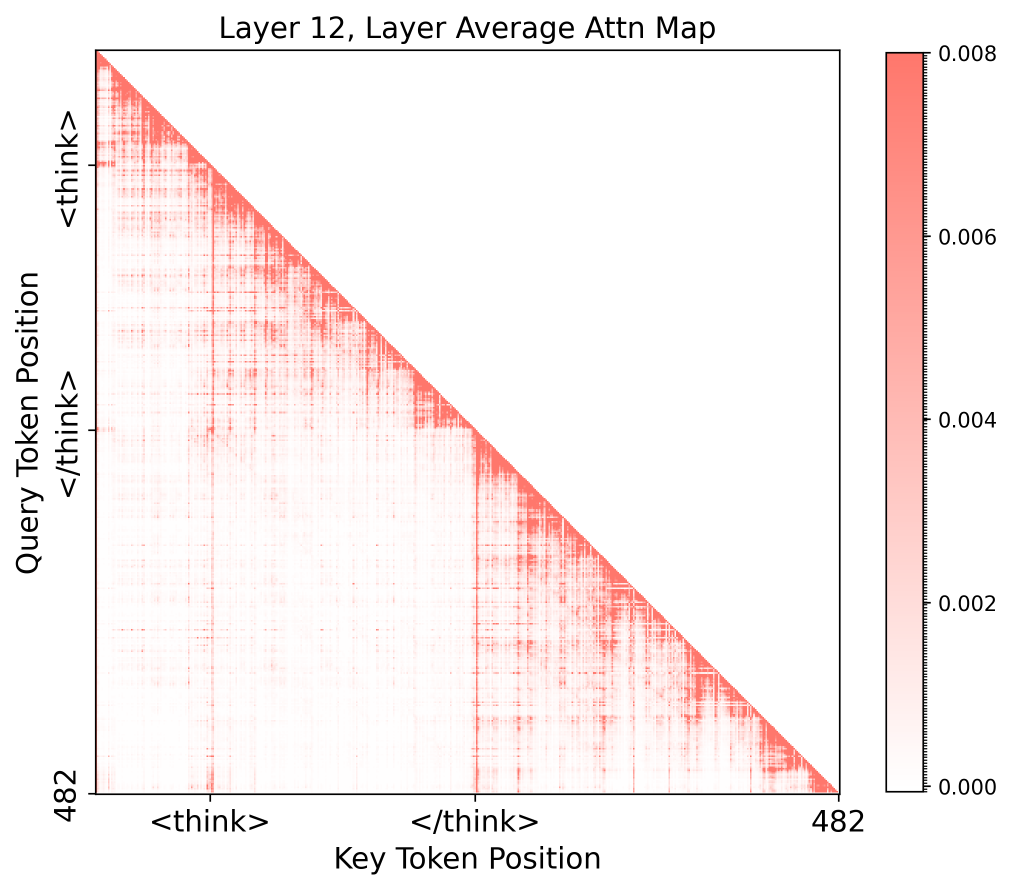}
        \caption{Layer 12}
    \end{subfigure}
    
    \vspace{0.5em}
    
    \begin{subfigure}{0.235\textwidth}
        \centering
        \includegraphics[width=\linewidth]{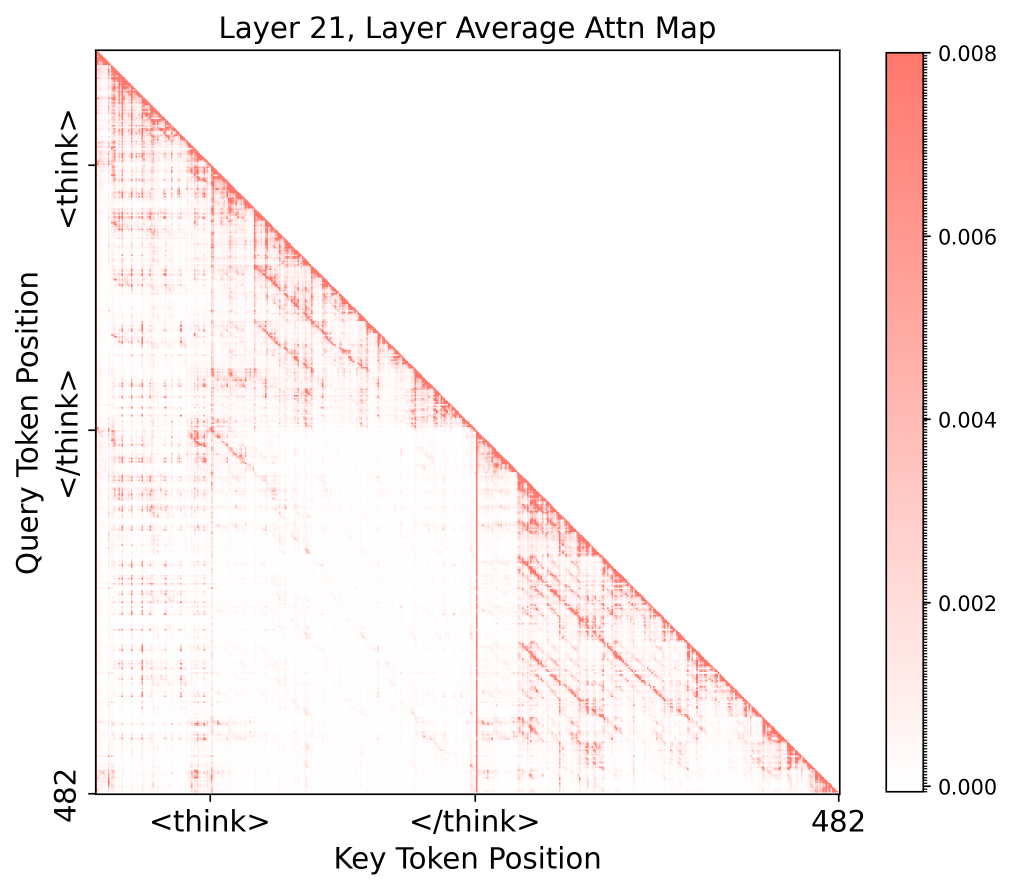}
        \caption{Layer 21}
    \end{subfigure}
    \hfill
    \begin{subfigure}{0.235\textwidth}
        \centering
        \includegraphics[width=\linewidth]{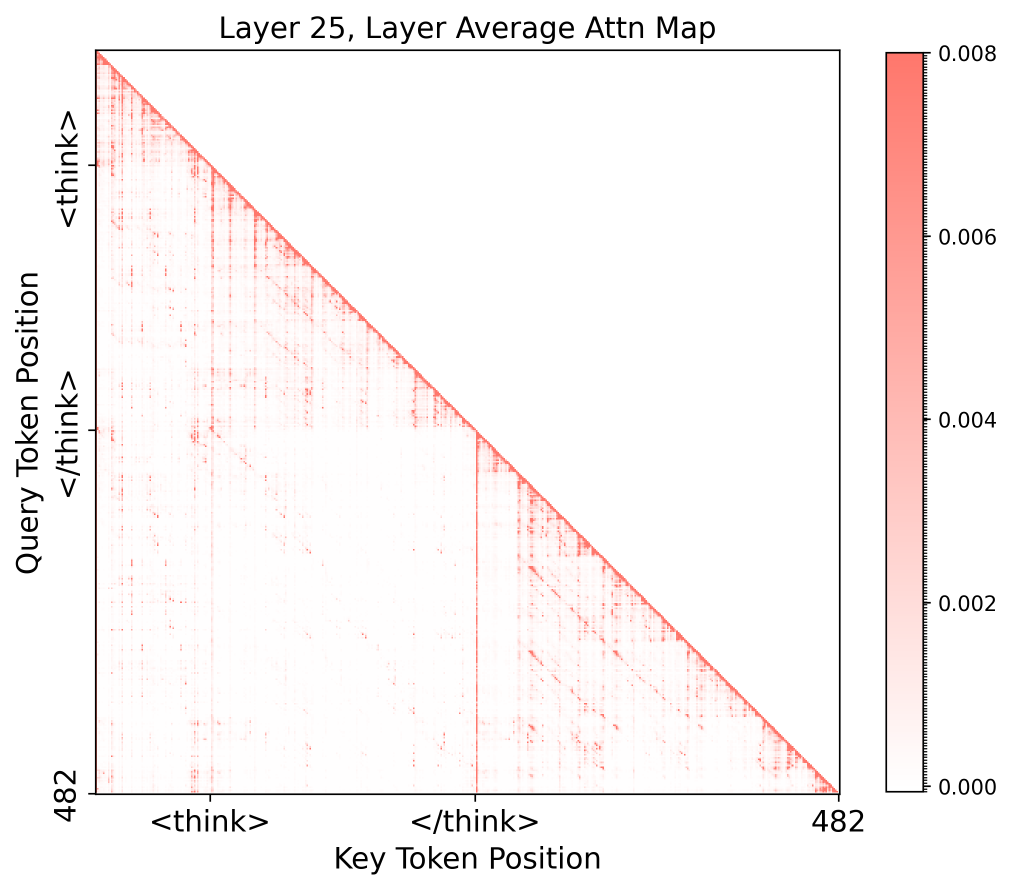}
        \caption{Layer 25}
    \end{subfigure}
    \hfill
    \begin{subfigure}{0.235\textwidth}
        \centering
        \includegraphics[width=\linewidth]{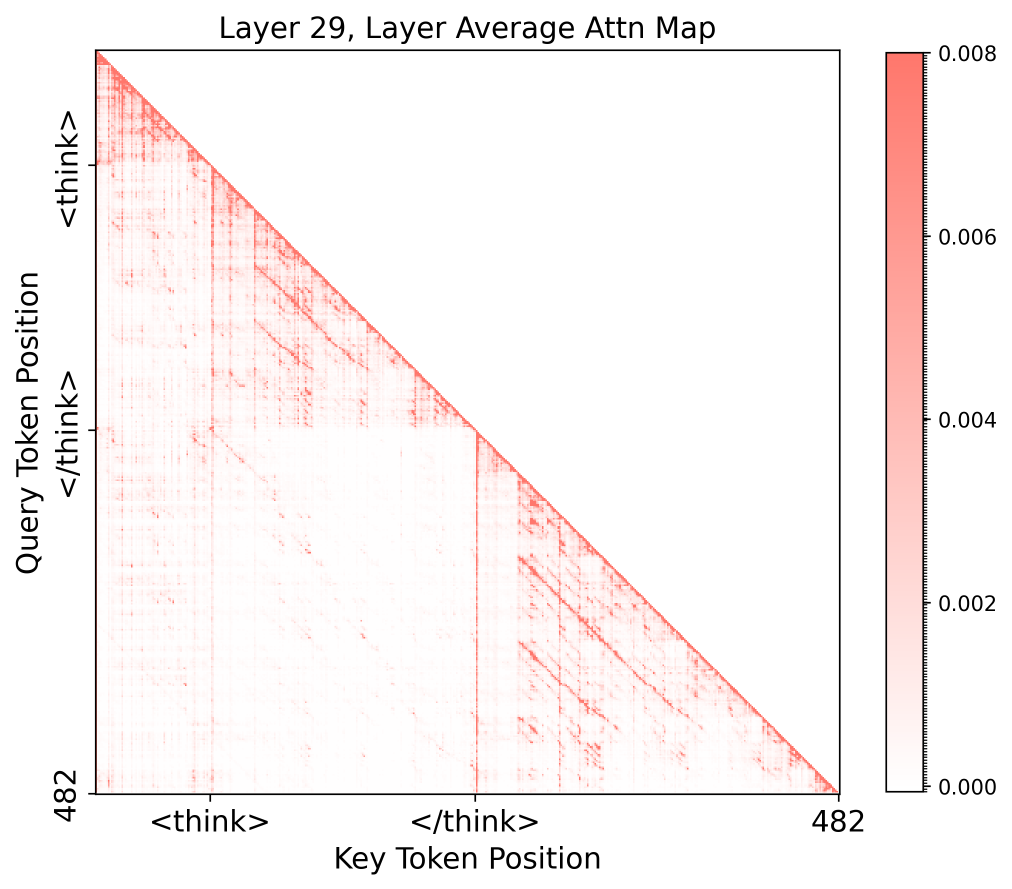}
        \caption{Layer 29}
    \end{subfigure}
    \hfill
    \begin{subfigure}{0.235\textwidth}
        \centering
        \includegraphics[width=\linewidth]{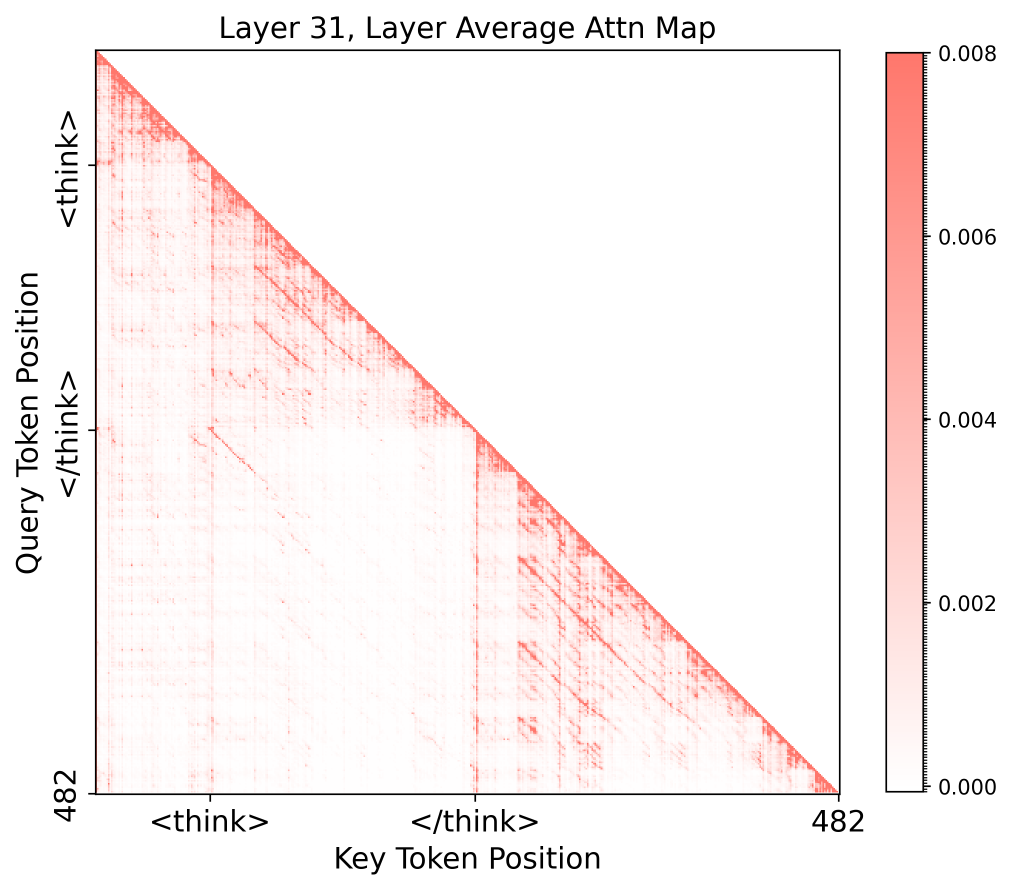}
        \caption{Layer 31}
    \end{subfigure}
    
    \caption{Attention heatmaps across different layers of DeepSeek-R1-Distill-LLaMA-8B on a GSM8K sample~\cite{cobbe2021training}. Tokens within the <$\mathrm{think}$>...<$\mathrm{/think}$> span receive uniform attention in early layers, but deeper layers gradually shift focus to the boundary tokens, indicating information migration and compression of reasoning content. Similar observations can be found in other models and datasets
    }
    \label{fig:attention_maps}
\end{figure*}

To understand why long-form CoT reasoning incurs high cost but limited benefit, we analzye the model's attention behavior during answer generation.
Our goal is to examine whether all reasoning tokens are equally useful—or if, as we hypothesize, later reasoning tokens alone may carry the necessary information for generating accurate answers.

We visualize attention patterns across transformer layers using DeepSeek-R1-Distill-Llama-8B on GSM8K samples, as shown in Figure\,\ref{fig:attention_maps}.

Each heatmap represents the attention weights from query tokens (rows) to key tokens (columns) during autoregressive decoding. The <$\mathrm{think}$> and <$\mathrm{/think}$> tokens mark the boundaries of the reasoning span. In early layers, the model distributes attention broadly across the reasoning region, suggesting that its initially considers the full reasoning race. However, as depth increases, the model's focus sharpens toward the end-of-reasoning boundary, particularly the <$\mathrm{/think}$> token. This transition implies a progressive information migration phenomenon, where reasoning content is gradually compressed toward the end of the span.

We attribute this behavior to causal masking: under left-to-right generation, downstream tokens cannot access future context. As results, reasoning must be internally summarized and propagated forward token by token. This leads earlier reasoning tokens to fade from view, while later tokens—particularly <$\mathrm{/think}$>—accumulate and represent the distilled reasoning state. Similar phenomena have been explored by~\cite{lin2025boosting}.

\begin{figure}[!t]
    \centering
    
    \begin{subfigure}{0.15\textwidth}
        \centering
        \includegraphics[width=\linewidth]{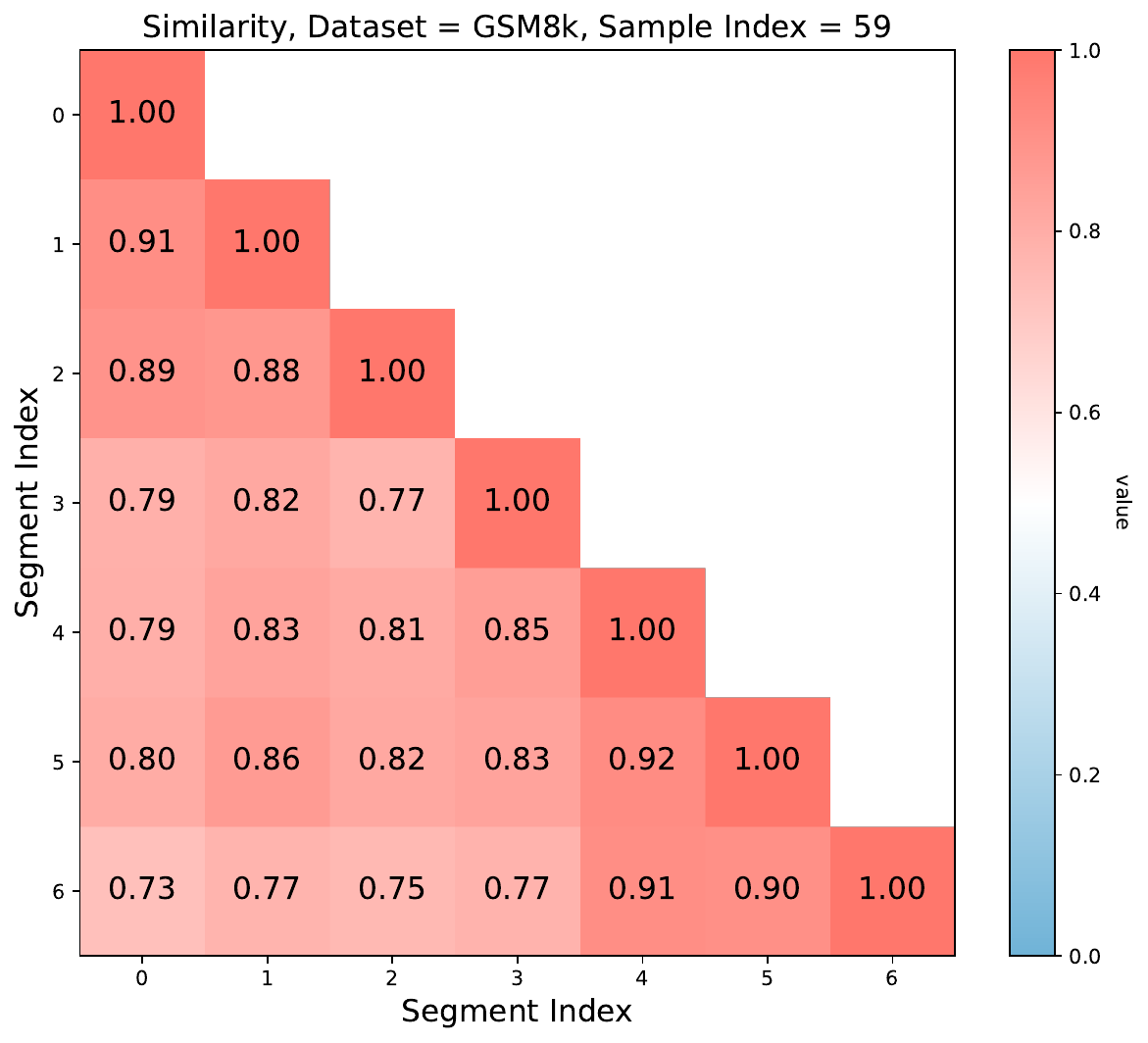}
        \caption{{\scriptsize Sample idx = 59}}
    \end{subfigure}
    \hfill
    \begin{subfigure}{0.15\textwidth}
        \centering
        \includegraphics[width=\linewidth]{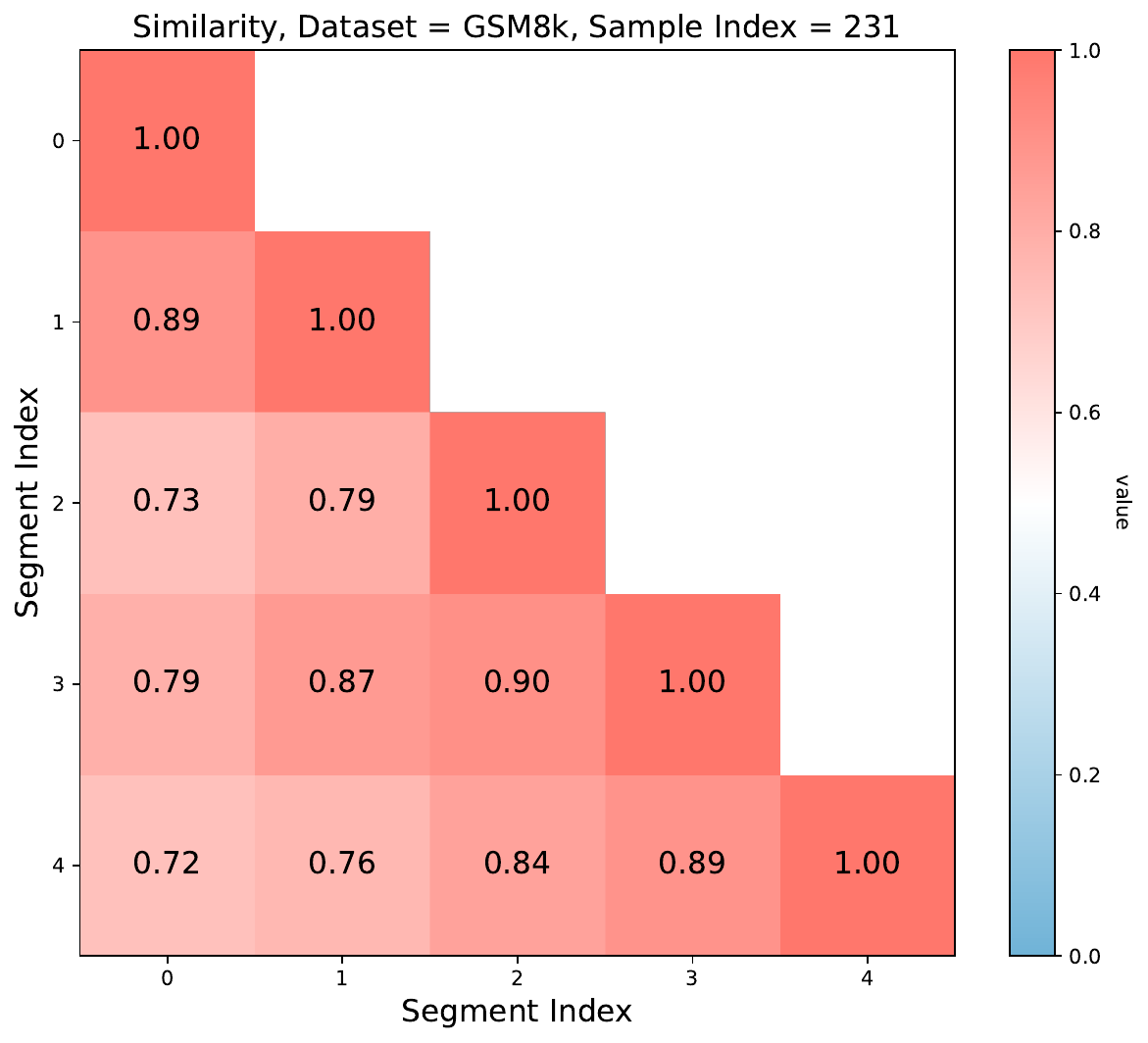}
        \caption{{\scriptsize Sample idx = 231}}
    \end{subfigure}
    \hfill
    \begin{subfigure}{0.15\textwidth}
        \centering
        \includegraphics[width=\linewidth]{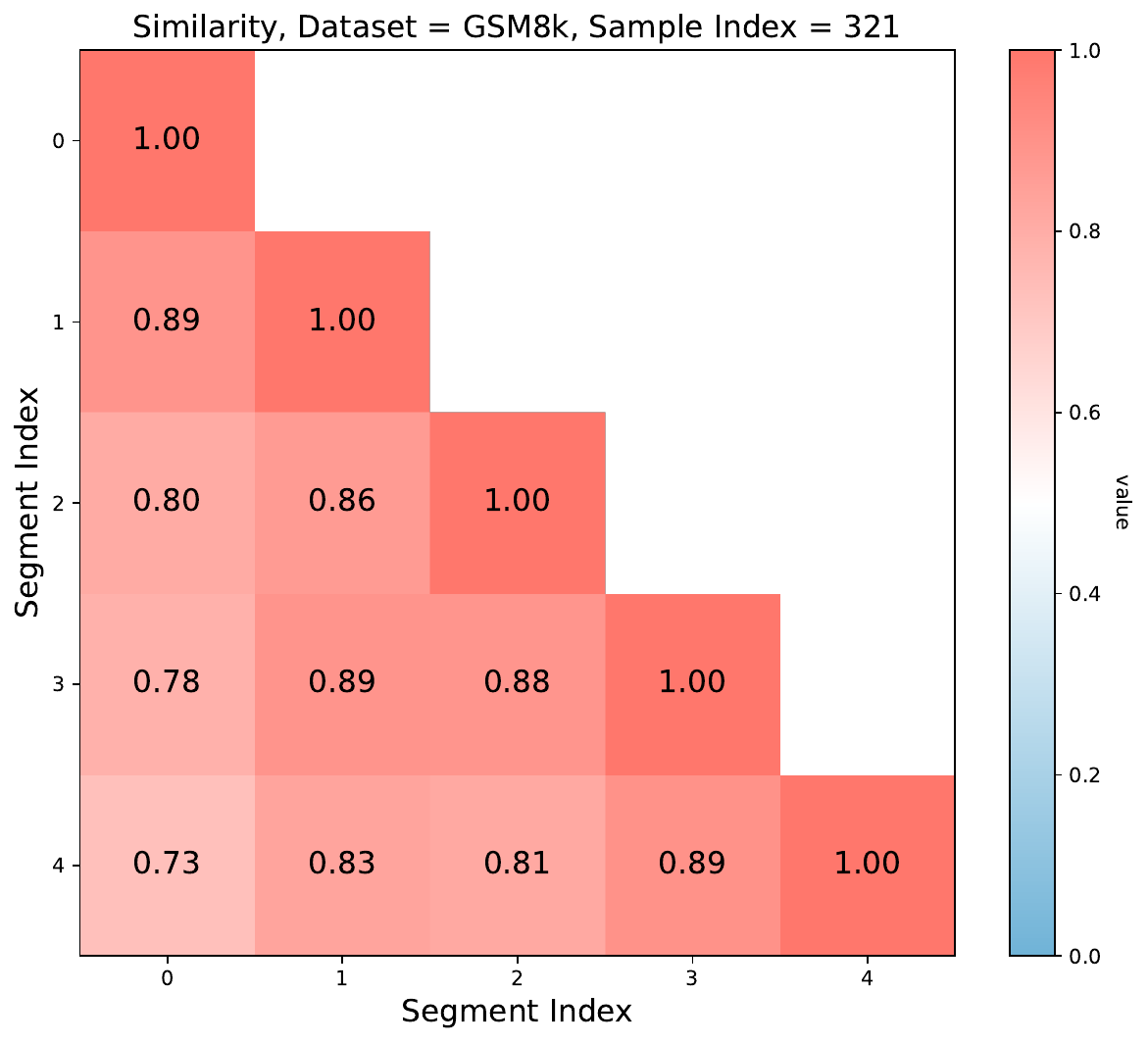}
        \caption{{\scriptsize Sample idx = 321}}
    \end{subfigure}  
    
    \begin{subfigure}{0.15\textwidth}
        \centering
        \includegraphics[width=\linewidth]{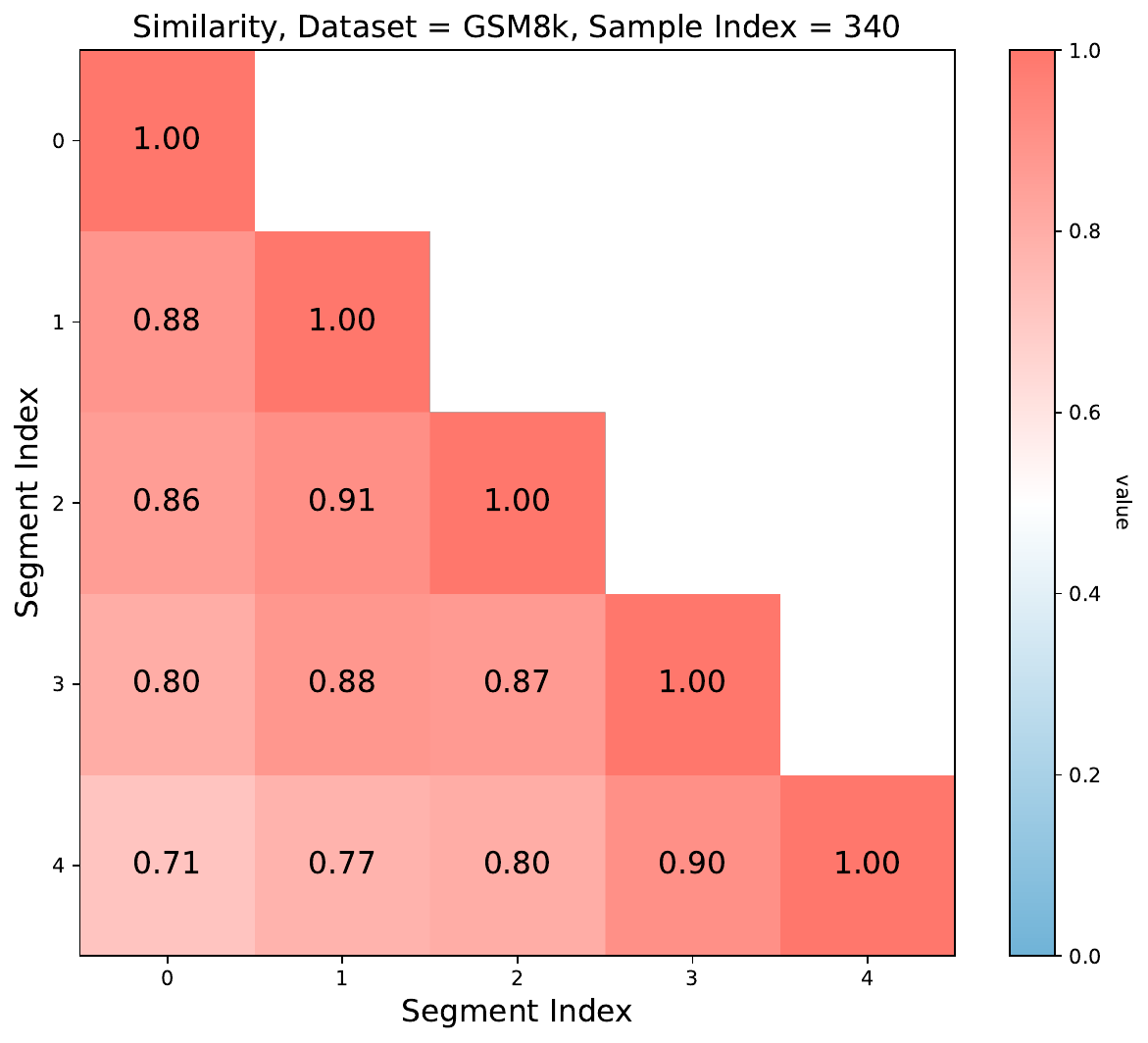}
        \caption{{\scriptsize Sample idx = 340}}
    \end{subfigure}
    \hfill
    \begin{subfigure}{0.15\textwidth}
        \centering
        \includegraphics[width=\linewidth]{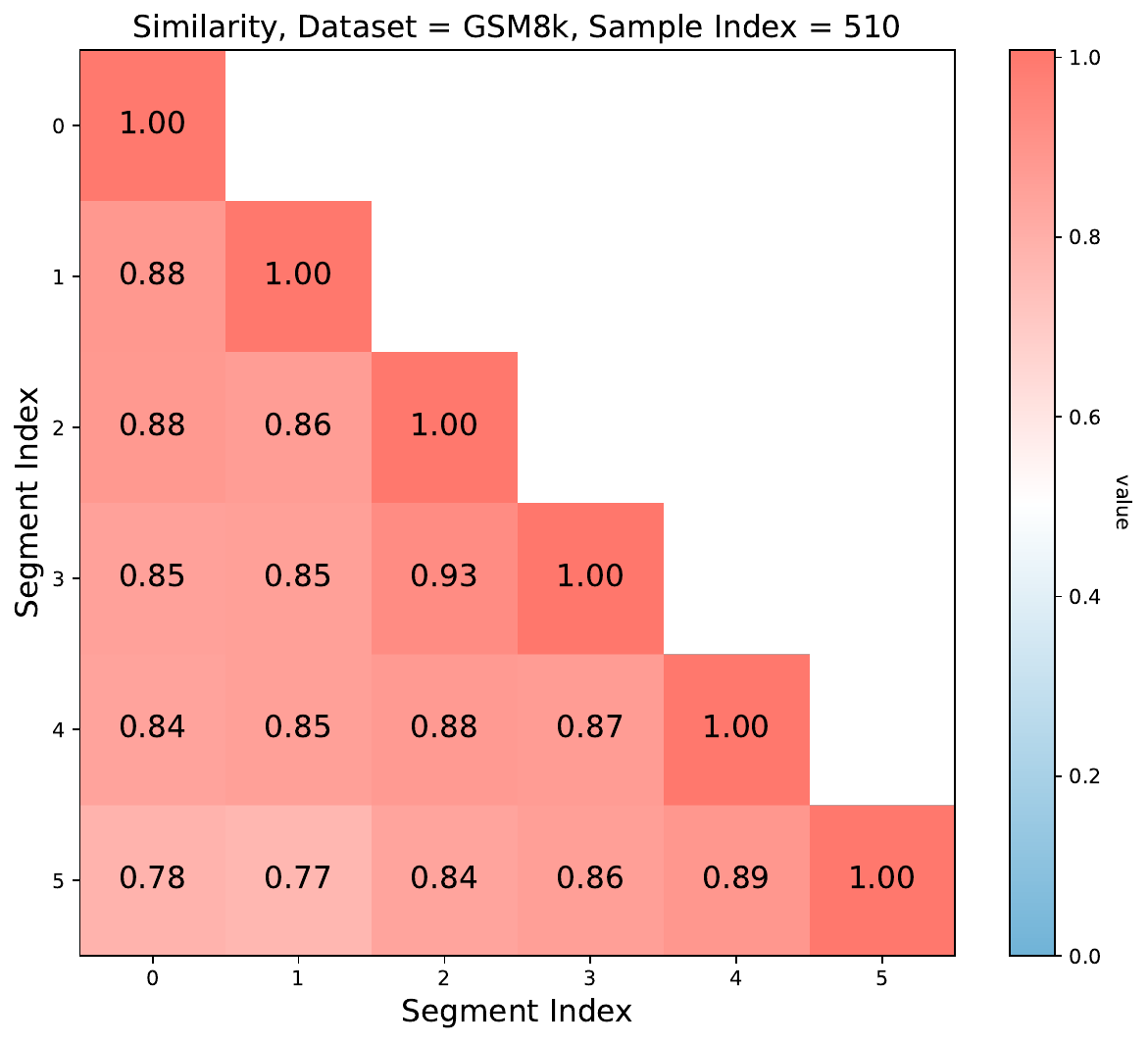}
        \caption{{\scriptsize Sample idx = 510}}
    \end{subfigure}
    \hfill
    \begin{subfigure}{0.15\textwidth}
        \centering
        \includegraphics[width=\linewidth]{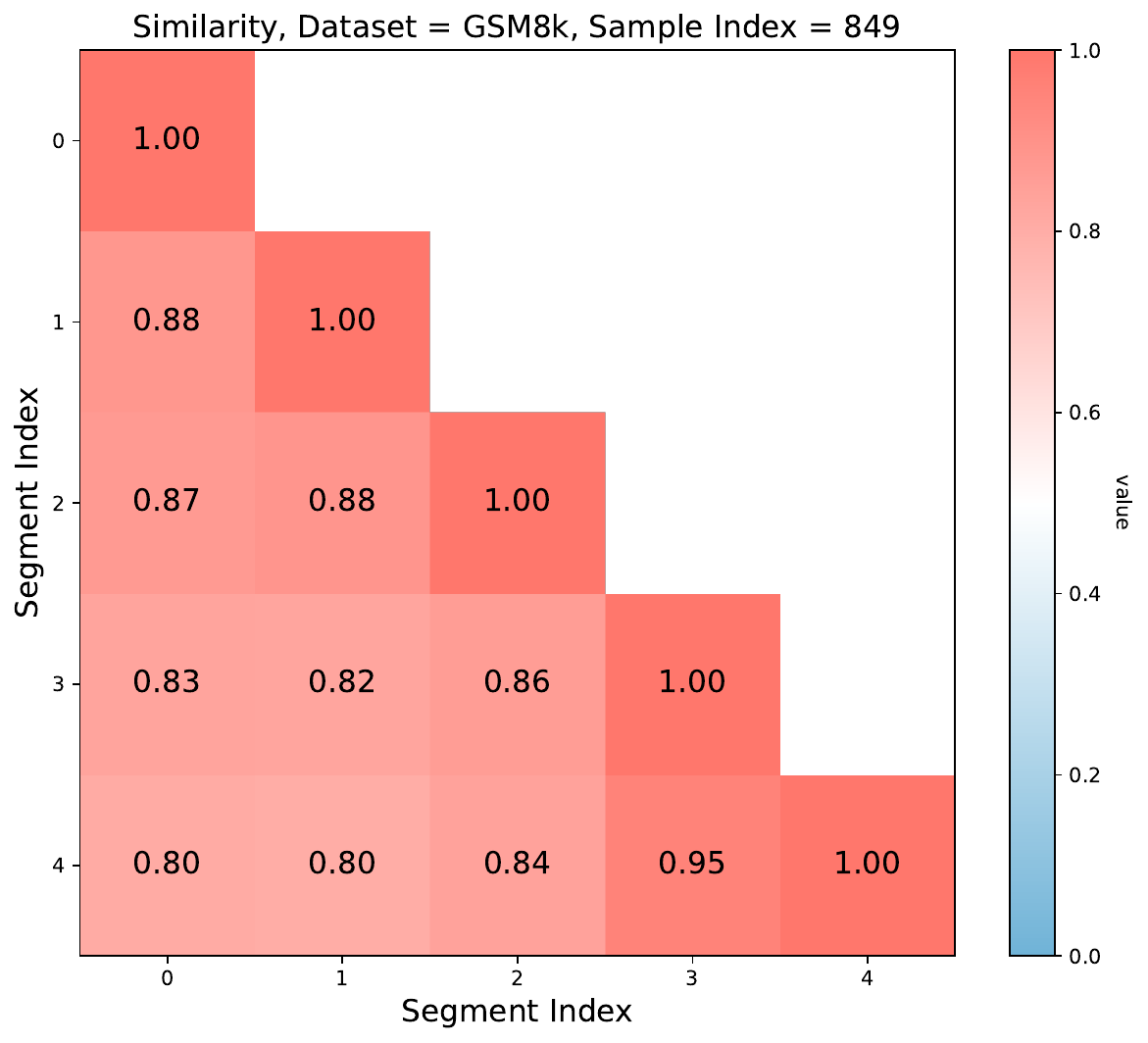}
        \caption{{\scriptsize Sample idx = 849}}
    \end{subfigure}
    
    \caption{We insert a <$\mathrm{/think}$> token every 16 tokens in DeepSeek-R1-Distill-Qwen-7B and extract last-layer hidden states. These states are highly similar (0.9) across segments, showing that reasoning adds little new information. The final state is also similar to earlier ones, indicating early convergence and redundancy in later reasoning. Similar observations can be found across other models and datasets. Best view with zooming in.}

    \label{fig:similarity_map}
\end{figure}

\paragraph{Analyzing Reasoning Redundancy.}
Building on the information migrration mechanism discussed above, we ask: \textit{How early can useful reasoning be distilled during generation?} Since reasoning content is expected to progressively compress toward the end of the span (\emph{e.g.}, <$\mathrm{/think}$>), we hypothesize that inserting this token at intermediate positions during decoding should yield hidden states that already approximate the final reasoning state. If true, this would suggest that the model has already internalized most of the reasoning content before completing the full chain.

To test this hypothesis, we conduct a similarity-based redundancy analysis. Specifically, we insert the <$\mathrm{/think}$> token at a fixed segment length of 16 tokens during the reasoning generation process using DeepSeek-R1-Distill-Qwen-7B. At each insertion point, we extract the last-layer hidden state of the <$\mathrm{/think}$> token, treating it as the representation of accumulated reasoning up to that step. We then compute pairwise cosine similarities between these intermediate hidden states.

As shown in Figure\,\ref{fig:similarity_map}, the similarity between adjacent reasoning segments remains consistently high ($\sim$0.9), indicating that each additional segment introduces only marginal new information. Moreover, the similarity between the final <$\mathrm{/think}$> state and earlier ones increases steadily, confirming the progressive nature of reasoning aggregation. Notably, even early inserted <$\mathrm{/think}$> tokens already yield hidden states highly similar to the final one—supporting the view that most useful reasoning content is distilled early, and extended CoT traces incur diminishing returns.

\begin{figure}[!t]
  \includegraphics[width=\linewidth]{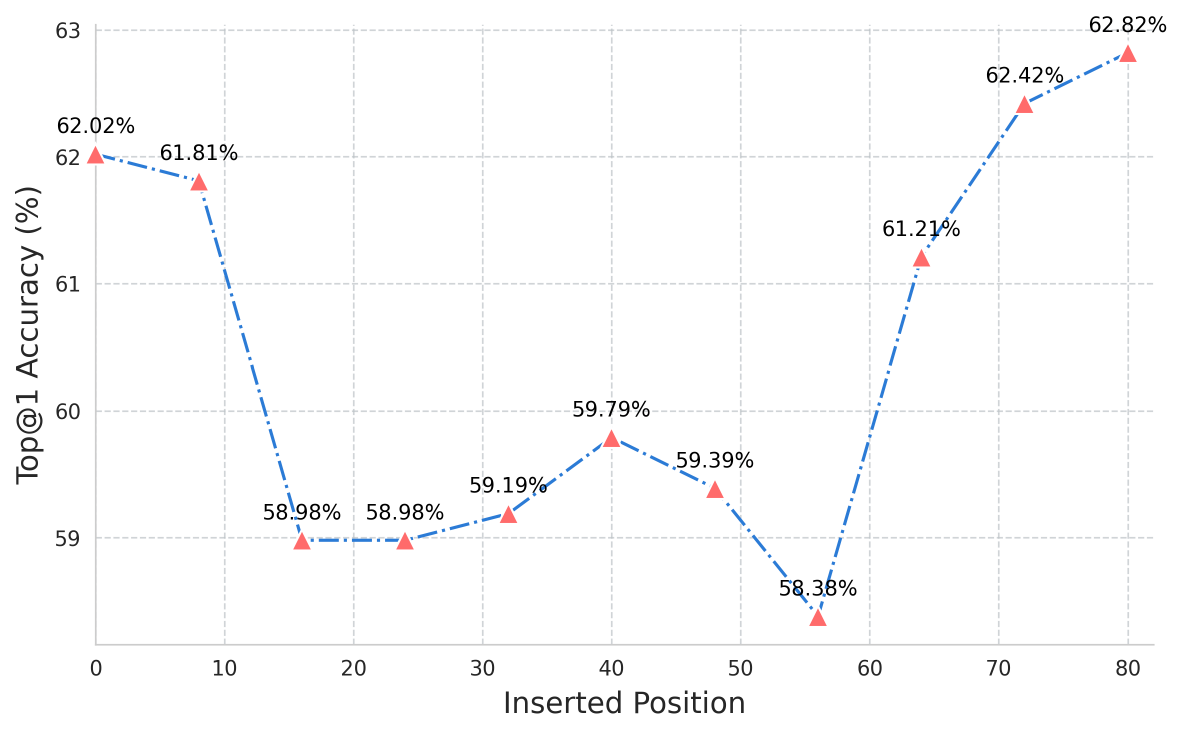}
  \caption{Accuracy of DeepSeek-R1-Distill-Qwen-7B \emph{vs.} position where <$\mathrm{/think}$> is inserted. The benchmark is BBH dataset~\cite{suzgun2022challenging}.}
  \label{fig:u-shaped}
\end{figure}

\subsection{ThinkLess: Reasoning Termination and Output Regulation\label{subsec:solution}}

Building on our earlier conclusion that the most useful reasoning content is distilled early, we ask: \textit{can reasoning be safely truncated early without sacrificing answer quality?} Since the model gradually compresses reasoning into the <$\mathrm{/think}$> token, it may be possible to shorten the reasoning trace while still preserving essential information.

To verify this, we divide the full reasoning sequence into equal-length segments and insert the <$\mathrm{/think}$> token at varying cut-off points, thereby terminating reasoning at different locations. We then measure model accuracy across termination positions. Surprisingly, as shown in Figure\,\ref{fig:u-shaped}, truncating reasoning early leads to a decreasing accuracy—despite our hypothesis that essential information should have already migrated toward the <$\mathrm{/think}$> token. This unexpected decline gradually recovers as the termination point moves later, forming a U-shaped performance curve.

A more detailed investigation in Sec.\,\ref{app:case} shows that the observed decline in performance is not attributable to deficiencies in the model's reasoning process. Instead, the drop primarily arises from output formatting issues—such as the omission of the final answer or deviations from the expected response structure. These formatting errors can lead to incorrect evaluations, even when the model's internal reasoning is logically sound. Notably, after manually correcting these malformed outputs to align with the desired answer format, we find the underlying responses are indeed accurate, resulting in a substantial recovery in overall accuracy.

This confirms that the observed accuracy dip is a surface-level artifact: early termination disrupts output form, not semantic correctness. The model had already internalized the reasoning; it simply failed to express it in the expected format.

These results confirm that the model primarily relies on the <$\mathrm{/think}$> token to access reasoning information—rather than attending to every reasoning token individually. As a result, extending the reasoning span offers limited benefit, revealing substantial redundancy in long-form CoT.

\paragraph{ThinkLess Framework.} We introduce \textbf{ThinkLess}, a simple, training-free framework to reduce CoT inference cost. The key idea is to insert the <$\mathrm{/think}$> token shortly after <$\mathrm{think}$>, thereby skipping the majority of reasoning generation. This early termination substantially reduces decoding time and KV cache memory usage. However, such abrupt truncation may produce malformed answers that lack structural completeness.

To overcome this dilemma, ThinkLess employs a lightweight instruction-based output regulation step. For each task, we prepend a short instruction prompt (see Sec.\,\ref{sec:instruction}) to clarify output expectations. This approach leverages the strong instruction-following abilities of modern LLMs, enabling the model to produce well-structured responses—even in the absence of explicit reasoning. Since the added instruction is minimal, the overall inference cost remains low.

\paragraph{Clarification: ThinkLess Without Explicit Reasoning.}
ThinkLess inserts the <$\mathrm{/think}$> token right after <$\mathrm{think}$>, thereby skipping the generation of any explicit CoT reasoning. At first glance, this appears to challenge the \textit{information migration hypothesis}: if no intermediate reasoning tokens are produced, it is unclear what reasoning content, if any, is being transferred to inform the final answer.

We contend, however, that the <$\mathrm{/think}$> token serves a deeper function than a mere delimiter. It acts as a \textit{semantic anchor}—a learned symbolic abstraction that implicitly encodes a compressed representation of the reasoning process. During pretraining, language models likely acquire the ability to internalize multi-step reasoning patterns and embed this abstracted knowledge into compact markers such as <$\mathrm{/think}$>. This hypothesis is supported by our empirical observations: even when the reasoning trace is entirely omitted, the model frequently produces correct answers, indicating that the cognitive process of reasoning may have been executed internally and silently.

From this perspective, <$\mathrm{/think}$> does not denote the absence of reasoning, but rather the culmination of an \textit{internalized} reasoning trajectory. It signals to the model that deliberation has concluded and that it should proceed to answer generation. This behavior can be interpreted as a form of \textit{reasoning distillation}, in which the explicit explanatory steps are compressed into latent activations, allowing for both efficient inference and high-quality outputs without requiring full CoT generation.

\section{Experiment}

\subsection{Datasets}\label{sec:datasets}
To comprehensively evaluate our proposed method across diverse reasoning and knowledge-intensive scenarios, we conduct experiments on the following four benchmark datasets:
\begin{itemize}
\item \textbf{GSM8K}~\cite{cobbe2021training}: A grade-school level math word problem dataset to assess arithmetic reasoning. Each question requires multi-step calculation and logical deduction.

\item \textbf{MMLU}~\cite{hendrycks2020measuring}: It covers 57 tasks across various domains including humanities, STEM, and social sciences, measuring general knowledge and reasoning ability.

\item \textbf{GPQA}~\cite{rein2024gpqa}: A graduate-level physics question answering dataset targeting conceptual understanding. It tests model capability in high-level scientific reasoning.

\item \textbf{BBH}~\cite{suzgun2022challenging}: This subset focuses on difficult tasks that require multi-step, symbolic, or logical reasoning, offering a rigorous stress test for language models.

\end{itemize}

\subsection{Metrics}

We report three key evaluation metrics across all tasks to provide a comprehensive comparison of both performance and efficiency:  Top-1 accuracy (Top@$1$$\uparrow$), inference time (Time$\downarrow$), and token usage (Tokens$\downarrow$). Accuracy reflects the percentage of exact top-1 matches. All results are from a single run.

Given that ThinkLess omits the explicit reasoning, we also report Top-$k$ accuracy (Top@$k$) ($k \ge 2$) for ThinkLess variants. In this setup, the model is allowed to generate $k$ candidate answers for each question, and the response is considered correct if any of them is accurate. This allows us to assess ThinkLess under a relaxed evaluation regime, which reflects its ability to retain answer quality even when reasoning tokens are suppressed.

To ensure fair comparison, we constrain the total number of generated tokens in the Top@$k$ setting to remain comparable to the token budget used by standard CoT decoding (\emph{i.e.}, Top@$1$ with full reasoning). This enables an apples-to-apples evaluation of accuracy under equivalent costs of tokens.

\subsection{Backbones and Baselines}
To ensure a comprehensive and fair evaluation, we conduct experiments on publicly available LLMs within the 7B to 14B parameter scale. This range reflects the practical constraints imposed by our available GPU resources, while still covering models with strong reasoning capabilities.

\textbf{Backbone.}
\textit{Qwen2.5-7B/14B}~\cite{yang2024qwen2}: A family of powerful open-source instruction-tuned models known for their strong general reasoning abilities.
\textit{LLaMA3.1-8B}~\cite{grattafiori2024llama}: A well-balanced model from the LLaMA series that combines efficient inference with competitive instruction-following performance.
All backbones are evaluated under identical decoding settings to ensure a consistent comparison.

\begin{table*}[!t]
    \centering
    \resizebox{\textwidth}{!}{
        \begin{tabular}{lccc|ccc|ccc|ccc|ccc}
            \toprule
            \multirow{2}{*}{\textbf{Method}} & \multicolumn{3}{c}{\textbf{GSM8K}} & \multicolumn{3}{c}{\textbf{MMLU}} & \multicolumn{3}{c}{\textbf{GPQA}} & \multicolumn{3}{c}{\textbf{BBH}} & \multicolumn{3}{c}{\textbf{AVG.}} \\
            \cmidrule(lr){2-4} \cmidrule(lr){5-7} \cmidrule(lr){8-10} \cmidrule(lr){11-13} \cmidrule(lr){14-16} 
            & \textbf{Top@$1\uparrow$} & \textbf{Time$\downarrow$} & \textbf{Tokens$\downarrow$} & \textbf{Top@$1\uparrow$} & \textbf{Time$\downarrow$} & \textbf{Tokens$\downarrow$} & \textbf{Top@$1\uparrow$} & \textbf{Time$\downarrow$} & \textbf{Tokens$\downarrow$} & \textbf{Top@$1\uparrow$} & \textbf{Time$\downarrow$} & \textbf{Tokens$\downarrow$}  & \textbf{Top@$1\uparrow$} & \textbf{Time$\downarrow$} & \textbf{Tokens$\downarrow$} \\
            \midrule
            \rowcolor{gray!15}\multicolumn{16}{c}{\textit{Qwen2.5-7B}} \\
            \midrule
            Distill & 88.17 & 10.62 & 438.92 & 60.86 & 47.01 & 1817.84 & 30.81 & 148.82 & 5523.17 & 69.29 & 24.79 & 976.08 & 62.28 & 57.81 & 2189.00\\
            ThinkLess w/o Instruct& 87.79 & 6.57 & 274.20 & 54.04 & 6.77 & 279.50 & 31.31 & 15.39 & 631.87 & 62.02 & 8.57 & 341.91 & 58.79 & 9.33 & 381.87\\
            ThinkLess & 88.40 & 5.46 & 235.41 & 57.06 & 9.07 & 370.34 & 40.91 & 14.59 & 591.17 & 65.25 & 9.34 & 379.32 & 62.91 & 9.62 & 394.06\\
            \midrule
            \rowcolor{gray!15}\multicolumn{16}{c}{\textit{Qwen2.5-14B}} \\
            \midrule
            Distill & 92.12 & 20.37 & 508.40 & 81.40 & 62.20 & 1516.46 & 41.92 & 217.62 & 5205.02 & 83.84 & 55.48 & 1349.88 & 74.82 & 88.92 & 2144.94\\
            ThinkLess w/o Instruct& 92.42 & 9.94 & 252.49 & 75.95 & 12.06 & 300.33 & 39.39 & 24.69 & 612.79 & 76.36 & 11.04 & 275.33 & 71.03 & 14.43 & 360.24 \\
            ThinkLess & 92.49 & 9.05 & 235.32 & 76.44 & 14.84 & 361.92 & 44.95 & 22.34 & 547.43 & 78.38 & 14.73 & 351.00 & 73.07 & 15.24 & 373.92 \\
            \midrule
            \rowcolor{gray!15}\multicolumn{16}{c}{\textit{LLaMA3.1-8B}} \\
            \midrule
            Distill & 79.38 & 12.95 & 493.70 & 64.07 & 56.69 & 2119.48 & 25.76 & 162.79 & 6094.77 & 71.92 & 33.21 & 1252.02 & 60.28 & 66.41 & 2489.99 \\
            ThinkLess w/o Instruct& 79.76 & 6.99 & 270.19 & 57.55 & 7.85 & 298.14 & 30.30 & 15.75 & 600.88 & 65.45 & 8.28 & 315.91 & 58.27 & 9.72 & 371.28 \\
            ThinkLess & 78.92 & 6.73 & 260.74 & 60.27 & 10.23 & 384.55 & 31.31 & 48.81 & 1817.93 & 71.92 & 11.45 & 430.89 & 60.61 & 19.31 & 723.53 \\
            \bottomrule
        \end{tabular}
    }
    \caption{Comparison of our ThinkLess and DeepSeek-R1 distilled models.}
    \label{tab:performance_comparison_g}
\end{table*}

\textbf{Baselines.}
We compare ThinkLess against a single, strong baseline: the full CoT distilled variant. This model is obtained by distilling reasoning capabilities from a more powerful DeepSeek-R1, and it represents a high-performance upper bound.

ThinkLess requires no fine-tuning, no auxiliary data, and no changes to the underlying model weights. To our best knowledge, we are the first to offer such efficient CoT reasoning compression in a fully training-free manner. Given this setting, the distilled full CoT model provides the most appropriate and meaningful baseline for comparison.
%

\subsection{How Effective is ThinkLess?}
\begin{figure*}[!t]

    \centering
    \begin{subfigure}{0.235\textwidth}
        \centering
        \includegraphics[width=\linewidth]{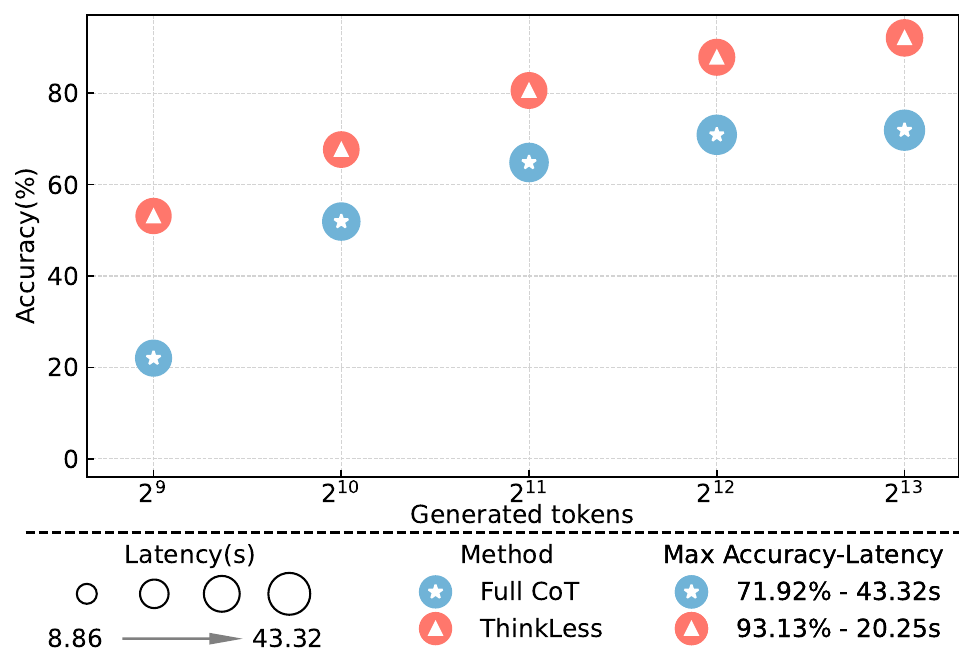}
        \caption{BBH, LLaMA}
    \end{subfigure}
    \hfill
    \begin{subfigure}{0.235\textwidth}
        \centering
        \includegraphics[width=\linewidth]{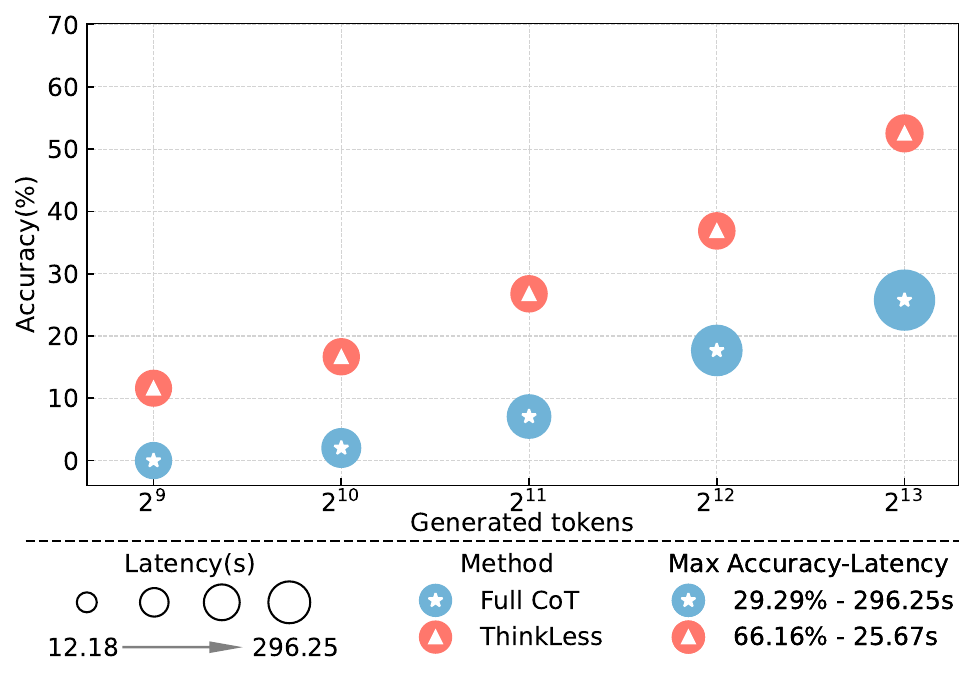}
        \caption{GPQA, LLaMA}
    \end{subfigure}
    \hfill
    \begin{subfigure}{0.235\textwidth}
        \centering
        \includegraphics[width=\linewidth]{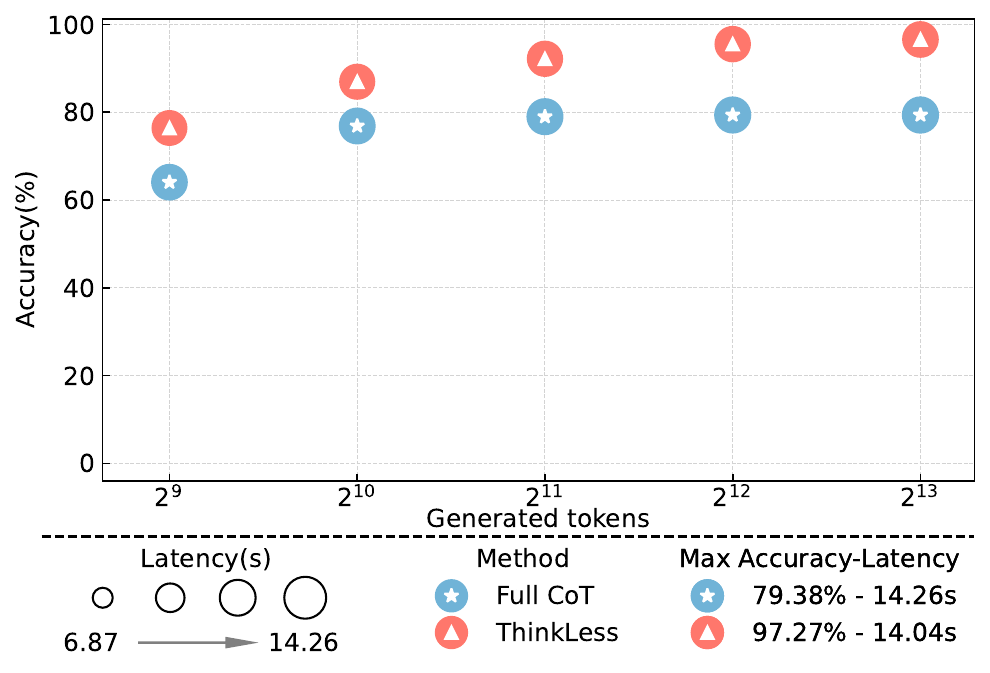}
        \caption{GSM8K, LLaMA}
    \end{subfigure}
    \hfill
    \begin{subfigure}{0.235\textwidth}
        \centering
        \includegraphics[width=\linewidth]{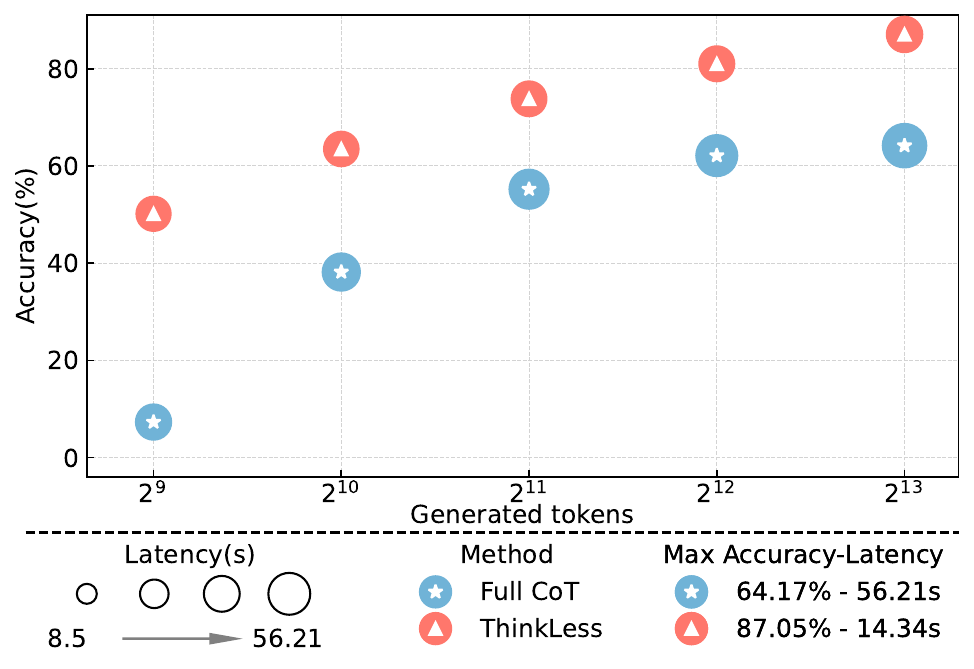}
        \caption{MMLU, LLaMA}
    \end{subfigure}
    
    \vspace{0.5em}
    
    \begin{subfigure}{0.235\textwidth}
        \centering
        \includegraphics[width=\linewidth]{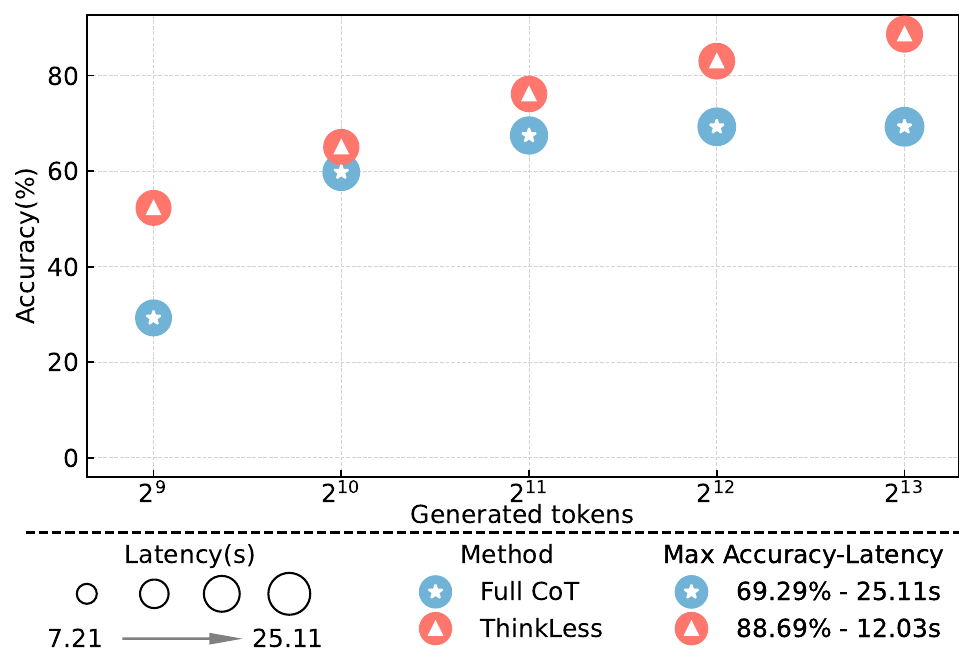}
        \caption{BBH, Qwen-7B}
    \end{subfigure}
    \hfill
    \begin{subfigure}{0.235\textwidth}
        \centering
        \includegraphics[width=\linewidth]{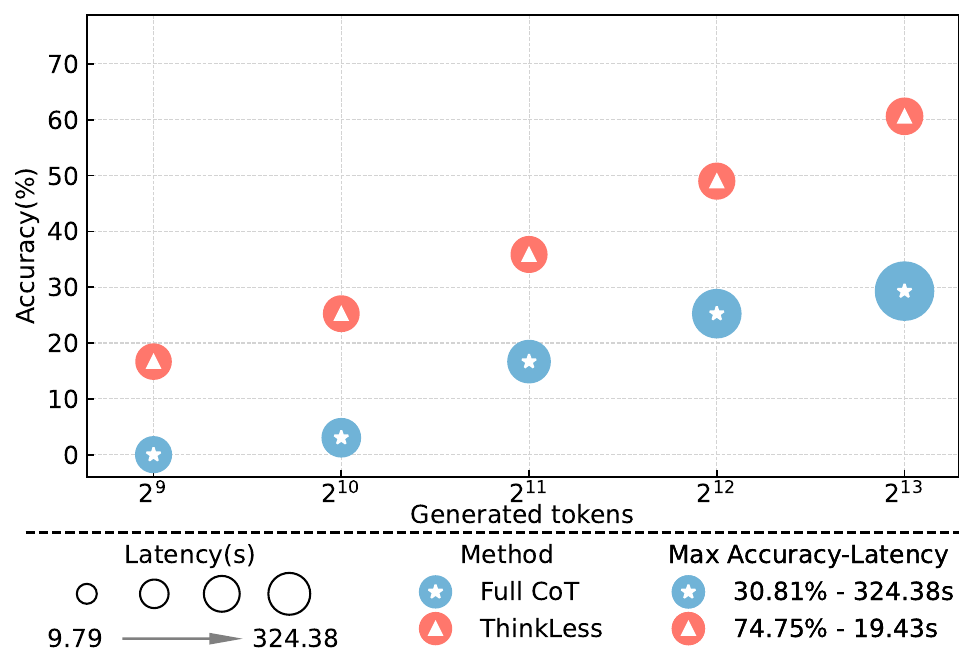}
        \caption{GPQA, Qwen-7B}
    \end{subfigure}
    \hfill
    \begin{subfigure}{0.235\textwidth}
        \centering
        \includegraphics[width=\linewidth]{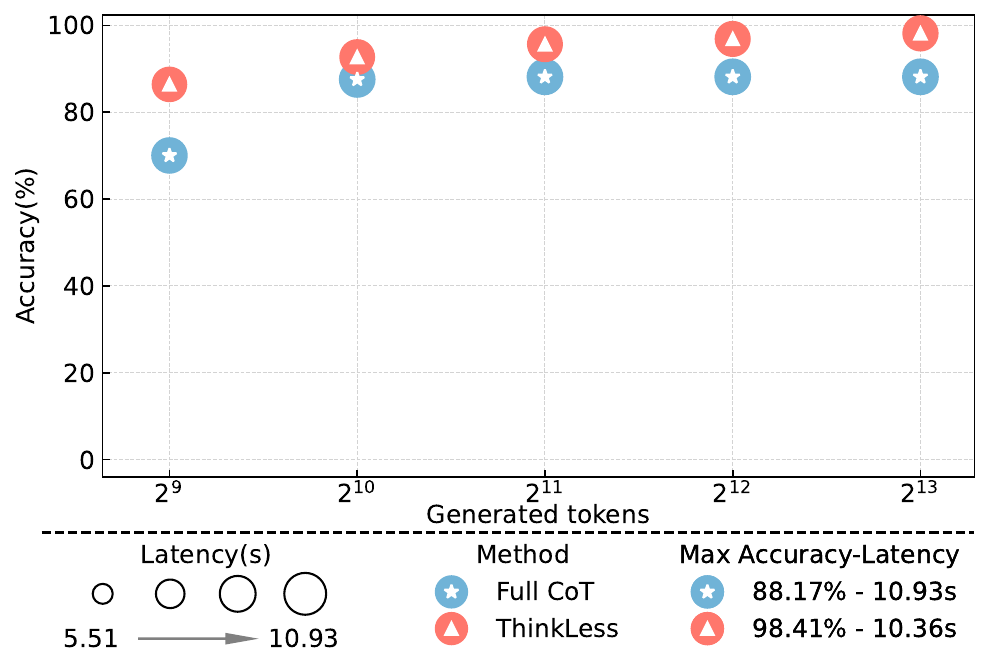}
        \caption{GSM8K, Qwen-7B}
    \end{subfigure}
    \hfill
    \begin{subfigure}{0.235\textwidth}
        \centering
        \includegraphics[width=\linewidth]{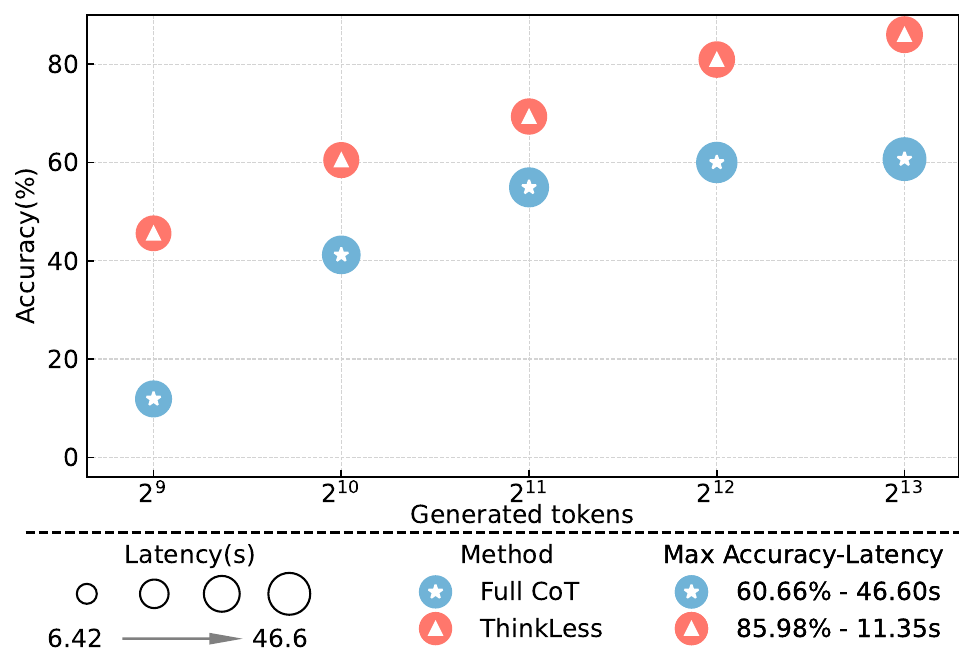}
        \caption{MMLU, Qwen-7B}
    \end{subfigure}

    \vspace{0.5em}
    
    \begin{subfigure}{0.235\textwidth}
        \centering
        \includegraphics[width=\linewidth]{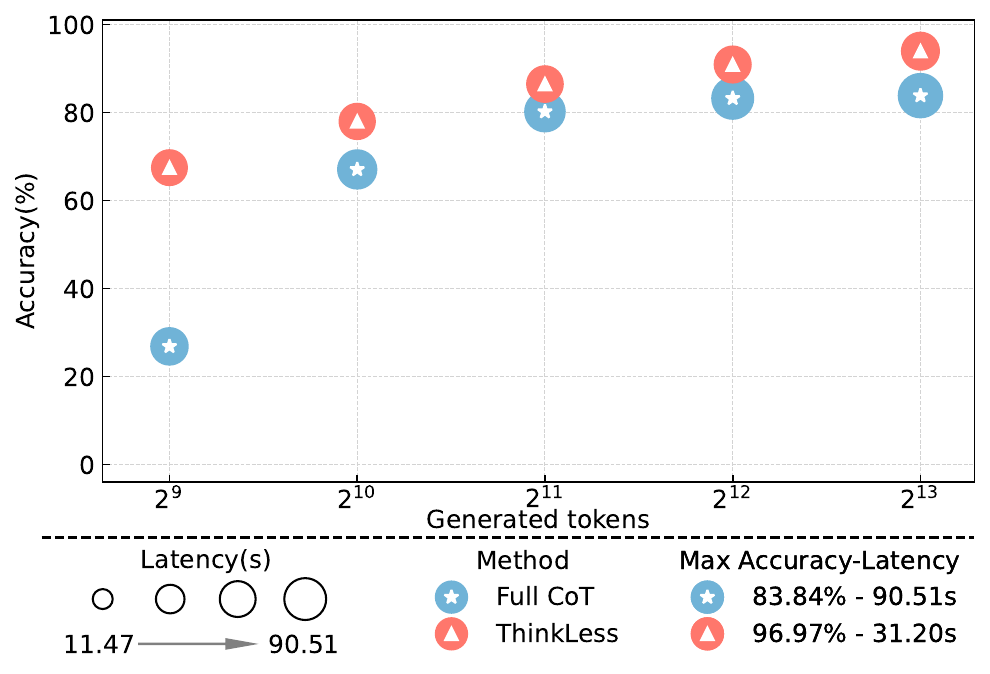}
        \caption{BBH, Qwen-14B}
    \end{subfigure}
    \hfill
    \begin{subfigure}{0.235\textwidth}
        \centering
        \includegraphics[width=\linewidth]{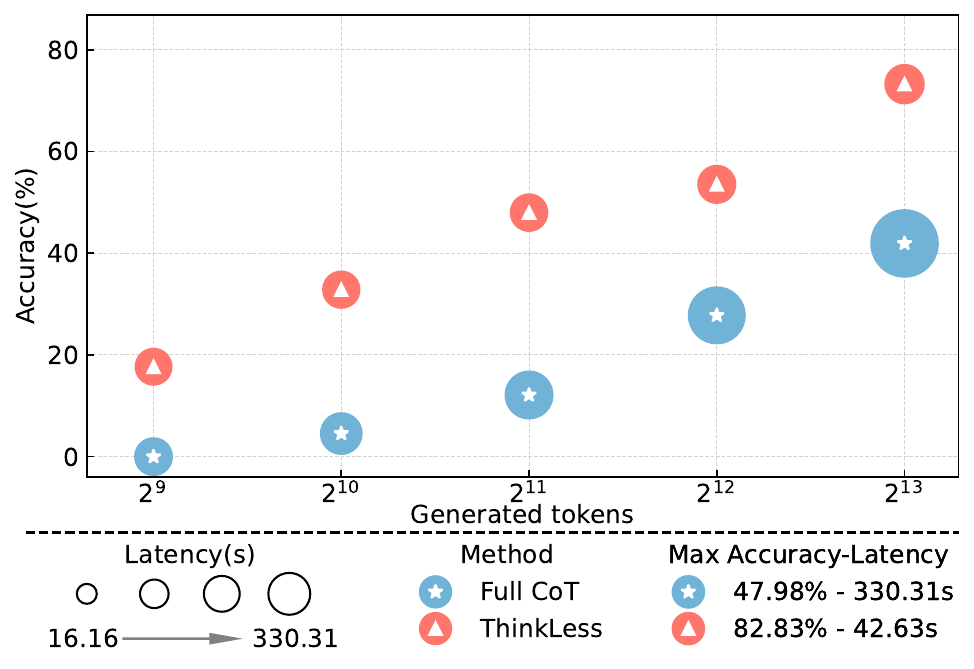}
        \caption{GPQA, Qwen-14B}
    \end{subfigure}
    \hfill
    \begin{subfigure}{0.235\textwidth}
        \centering
        \includegraphics[width=\linewidth]{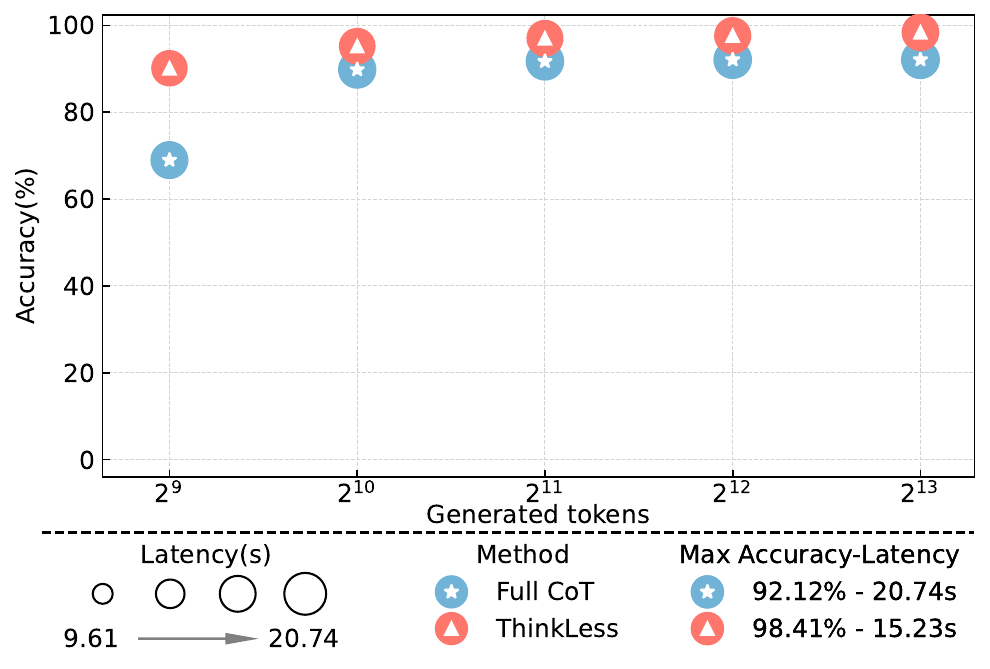}
        \caption{GSM8K, Qwen-14B}
    \end{subfigure}
    \hfill
    \begin{subfigure}{0.235\textwidth}
        \centering
        \includegraphics[width=\linewidth]{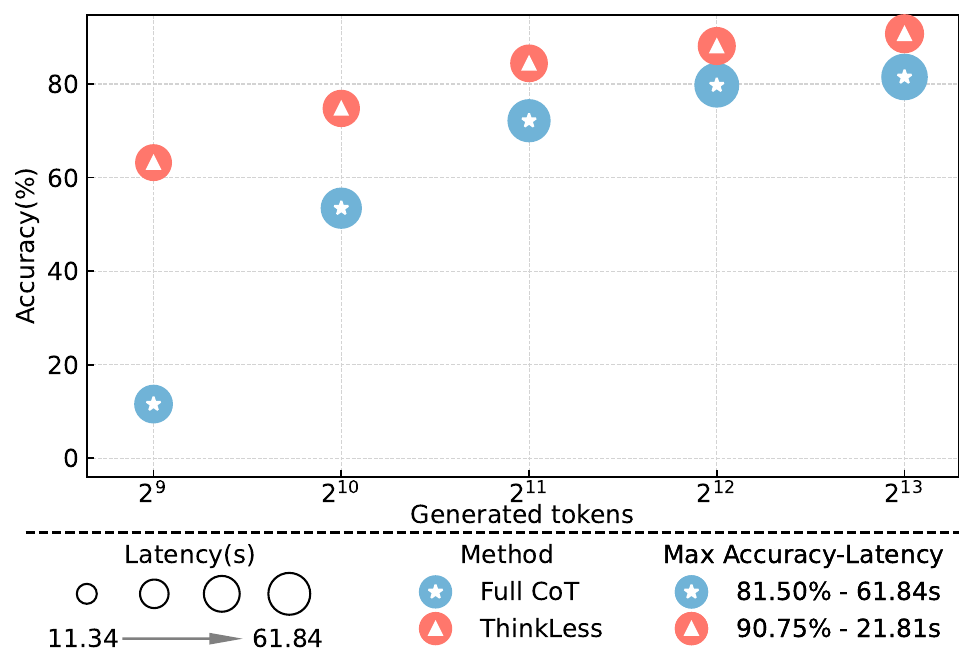}
        \caption{MMLU, Qwen-14B}
    \end{subfigure}

    \caption{{\small Top@$k$ accuracy of ThinkLess \emph{vs.} Top@$1$ accuracy of DeepSeek-distilled models across datasets and models. We set $k = \frac{\text{Token Budget}}{512}$ to match the token usage on par with distilled models. Legends follow Figure~\ref{fig:illustration}.}}
    \label{fig:same_budget}
\end{figure*}

Table\,\ref{tab:performance_comparison_g} and Figure\,\ref{fig:same_budget} present a detailed comparison between our proposed \textbf{ThinkLess} framework and the \textbf{Distill} baseline. The maximum token budget is set as 8k in Table\,\ref{tab:performance_comparison_g}. We detail accuracy, inference time, and token consumption below.

\textbf{Comparable Accuracy Despite Omitting Reasoning.}
While ThinkLess entirely skips the visible CoT reasoning trace, its Top@$1$ accuracy remains consistently close to that of the full CoT baseline. For example, with Qwen2.5-7B, ThinkLess achieves an average accuracy of 62.91\%, compared to 62.28\% from Distill. With Qwen2.5-14B, ThinkLess reaches 73.07\% \emph{vs}. 74.82\%. These small differences—within 1–2 points—demonstrate that ThinkLess retains most of the reasoning quality, validating our core hypothesis: reasoning can be effectively compressed into latent representations without explicit CoT generation.

\textbf{Enhanced Accuracy under Comparable Token Budgets.}
Figure\,\ref{fig:same_budget} presents the Top@$k$ accuracy of ThinkLess compared against the Top@1 accuracy of the full CoT Distill baseline, under an equal token budget. The results show that ThinkLess significantly outperforms the distilled counterpart across various datasets and model backbones. Notably, beyond accuracy improvements, ThinkLess also achieves lower inference latency. This is because the $k$ candidate answers in ThinkLess can be generated in parallel, whereas the distilled baseline must generate a long CoT sequence token by token in an inherently sequential manner.

\begin{figure*}[!t]
  \includegraphics[width=\linewidth]{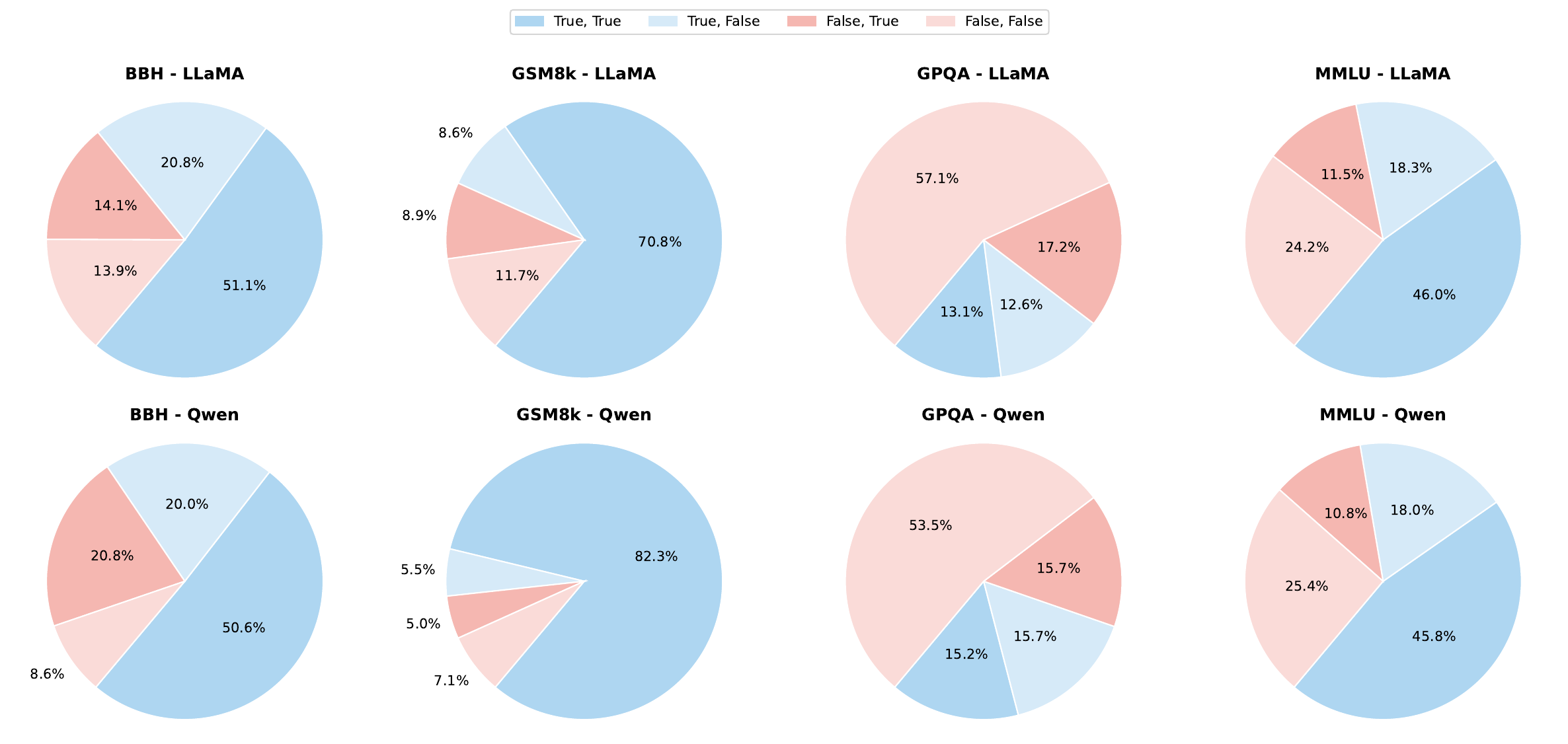}
  \caption{Answer overlap between Distill and ThinkLess w/o Instruct. Each pie shows the proportion of ``a, b'' cases, where ``a'' is Distill's results (True or False) and ``b'' is ThinkLess w/o Instruct's (True or False).}
  \label{fig:classification}
\end{figure*}

\textbf{Massive Reductions in Token Usage and Inference Time.}
ThinkLess achieves substantial efficiency gains across all settings. On average:
\textit{1. Token usage is reduced by 60–70\%}, dropping from 2189 tokens (Qwen2.5-7B, Distill) to just 904 with ThinkLess.
\textit{2. Inference time is reduced by 50\% or more}, \emph{e.g.}, from 21.89s to 9.64s with Qwen2.5-14B.
These savings stem from truncating long reasoning sequences early via <$\mathrm{/think}$>, which eliminates most of the token generation and KV cache accumulation that typically burdens autoregressive inference. Crucially, these gains come without any fine-tuning, distillation, or prompt engineering, making ThinkLess easy to deploy.

\textbf{Robustness Across Models and Tasks.}
Although ThinkLess occasionally underperforms on specific datasets (\emph{e.g.}, slightly lower on BBH with Qwen2.5-14B), its average accuracy is remarkably stable across all backbones. This consistency indicates that our method generalizes well across diverse reasoning tasks and model families.

\textbf{The Role of Output Regulation.}
Comparing ThinkLess to its ablated version \textbf{ThinkLess w/o Instruct} highlights the impact of our lightweight instruction-based output regulation. Across all settings, ThinkLess consistently outperforms the w/o Instruct variant in Top@1 accuracy, often by a significant margin. For instance:
On MMLU with Qwen2.5-14B: ThinkLess achieves 76.44\% \emph{vs.} 75.22\%.
On BBH with LLaMA3-8B: 71.92\% \emph{vs}. 65.45\%, a gap of over 6 points.

Figure\,\ref{fig:classification} illustrates the answer agreement between Distill and ThinkLess w/o Instruct across datasets and backbones. Across most of the datasets, over 70\% of predictions remain consistent (\emph{i.e.}, <True, True> or <False, False>), demonstrating ThinkLess can well preserve ability of the Distill model despite its early termination.

This confirms that without output regulation, the model—though internally sound—frequently fails to produce well-structured answers (\emph{e.g.}, missing final choice or wrong format). The addition of a short task-specific instruction guides the model to produce answers in a predictable and scorable format, which is critical for maintaining accuracy in the absence of full reasoning traces.

\textbf{Summary.}
ThinkLess achieves comparable Top@1 accuracy to full CoT reasoning while halving inference time and reducing token usage by up to 70\%, all in a training-free and model-agnostic manner. These results demonstrate that ThinkLess offers a highly practical trade-off between reasoning fidelity and computational efficiency.

\section{Conclusion}
This paper presents \textbf{ThinkLess}, an inference-efficient framework that reduces the overhead of CoT reasoning without any model modification or additional training. By analyzing attention patterns, we find that final answers rely little on early reasoning steps—enabling safe early termination via a reasoning terminator token. To preserve answer completeness and format, a lightweight output regulation step is introduced, leveraging the model’s instruction-following ability. Experimental Results show that ThinkLess achieves comparable accuracy to full CoT decoding while significantly lowering token usage and latency, making it a practical and generalizable solution for real-world deployment.

\section{Limitations}
While ThinkLess demonstrates strong efficiency and accuracy trade-offs, several limitations remain:

\textbf{Reliance on Instruction Quality.}
The success of ThinkLess depends on the effectiveness of lightweight output regulation instructions. Poorly phrased or overly generic instructions may fail to guide the model toward well-structured outputs, especially for complex or ambiguous tasks. Designing effective instructions for new tasks may require manual tuning or domain-specific insights.

\textbf{Lack of Dynamic Truncation Strategy.}
ThinkLess currently inserts the <$\mathrm{/think}$> token at fixed positions, without dynamically adapting to the complexity of individual questions. For harder tasks requiring deeper reasoning, premature truncation may omit essential content. Developing an adaptive termination policy that tailors reasoning length to question difficulty remains an open direction.

\textbf{Assumption of Internal Reasoning Compression.}
ThinkLess assumes that LLMs internally compress reasoning into the <$\mathrm{/think}$> token, which may hold for certain instruction-tuned models but not all. Models without strong instruction-following capabilities or those trained with different prompting formats may not benefit from early termination in the same way, limiting the generalizability of our method.

\textbf{Limited Scalability Validation.}
Due to computational resource constraints, we only evaluate ThinkLess on mid-sized models (7B–14B) and a limited set of reasoning benchmarks. Its performance on larger foundation models or broader tasks remains to be validated.

These limitations also highlight important directions for future work. In particular, extending ThinkLess to larger-scale models, more diverse task types, and dynamic truncation policies remains a key focus of our ongoing efforts.

\section{Ethical Considerations}

We use publicly available datasets and model checkpoints under licenses that permit research use. Details about the license terms and usage restrictions are provided in Section \ref{sec:datasets}.
We ensured that all artifacts were used in accordance with their intended purpose as stated by the original providers.


\bibliographystyle{acl_natbib}

\clearpage

\appendix

\clearpage

\section{Analysis of Output Formatting Issues from Early Termination\label{app:case}}

While ThinkLess is designed to terminate reasoning early and rely on internalized representations for answer generation, this can occasionally result in malformed outputs—particularly when the model is not explicitly instructed on how to format its final answer. Tables\,\ref{tab:gpqa_case} and \ref{tab:bbh_case} illustrate common failure cases across different datasets, caused not by flawed reasoning, but by formatting deviations that hinder correct evaluation.

\textbf{GPQA: Outputting the Answer Directly Instead of the Option.}
In multiple GPQA examples, the model correctly computes the numerical answer (\emph{e.g.}, ``18'', ``$\frac{1 + nv}{n + v}$'', or ``3536''), but fails to select the corresponding multiple-choice option letter (\emph{e.g.}, ``A'', ``B'', \emph{etc}.). This is problematic because the task requires choosing from a list, and direct numeric answers—though logically correct—are treated as incorrect under automatic evaluation scripts. This issue is a direct consequence of skipping the reasoning trace, which would otherwise reinforce the expected answer format (\emph{e.g.}, ``The answer is A'').

\textbf{BBH: Verbose or Misaligned Natural Language Outputs.}
In BBH, early termination sometimes causes the model to output full explanations (\emph{e.g.}, ``No, Tamika does not tell the truth'') instead of a concise boolean token like “False” or a lettered choice. In one example, the model responds with an overly verbose phrase: ``The statement `Return your final response within boxed \{\}' is True''—a hallucination likely caused by partial instruction remnants seen during pretraining. These cases reveal how early truncation may weaken task compliance, especially for boolean or classification-based tasks that expect minimal final output.

\textbf{MMLU: Misformatting Algebraic Expressions.}
For math-heavy tasks like MMLU, the model sometimes produces an exactly correct symbolic expression (\emph{e.g}., ``$(x + 1)(x - 2)(x + 4)$'') that does not match any of the provided answer options verbatim. Though mathematically equivalent to one of the choices, the mismatch in formatting or token order causes the model to be penalized. This highlights the fragility of matching-based evaluation when outputs are not explicitly aligned with options.

\textbf{Key Insight: Output Regulation is Essential.}
These examples demonstrate that output formatting errors—not reasoning failures—are the dominant cause of performance drop in ThinkLess without instruction-based regulation. The missing or misaligned final answers occur because the model lacks an explicit signal about how to conclude the response after <$\mathrm{/think}$> is triggered.

By contrast, ThinkLess with output regulation prepends a short, task-specific instruction (\emph{e.g.}, ``Select the best option (A, B, C, D):'') that helps the model map internal reasoning to a valid and scorable final output—without increasing token length significantly. This regulation mechanism is crucial for ensuring compatibility with automatic scorers and maintaining downstream performance.

\section{Instruction-based Output Regulation\label{sec:instruction}}

Tables\,\ref{tab:bbh} and \,\ref{tab:others} provides instructions details that regulate the output formatting across different datasets and their subtasks.

\begin{table*}[htbp]
\small
\renewcommand\arraystretch{1.3}
\centering
\begin{tabularx}{\textwidth}{c|X|X}
\hline
\textbf{Dataset} & \textbf{Input} & \textbf{Output} \\
\hline

\multirow{2}{*}{GPQA} 
& If an equimolar mixture X of two liquids, which decolorizes bromine water, is treated with platinum when heated, 
then an equimolar mixture Y of two other liquids is formed as a result of disproportionation, which does not decolorize bromine water. Hydrogenation of both mixture 
X and mixture Y in the presence of platinum under rather severe conditions gives only one substance, a certain hydrocarbon Z (mass fraction of hydrogen is 14.28\%),
 which is a constituent of mixture Y and widely used as a solvent. Substance Z does not react further with hydrogen. 
There are no conjugated multiple bonds in the molecules of the compounds of mixture X.
Indicate the total number of hydrogen atoms in two liquids of mixture X.
\begin{itemize}
    \item[A.] \( 18 \)
    \item[B.] \( 22 \)
    \item[C.] \( 16 \)
    \item[D.] \( 12 \)
\end{itemize} & 18 \\
\cline{2-3}
& A light beam is propagating through a glass with index of refraction n. The glass is moving at constant velocity v in the same direction as the beam and toward the observer in laboratory. What is the speed of light in glass relative to the observer in laboratory? Take the speed of light in vacuum c=1.
\begin{itemize}
    \item[A.] \( 1 \)
    \item[B.] \( \dfrac{1 - n v}{n + v} \)
    \item[C.] \( \dfrac{1 + n v}{n + v} \)
    \item[D.] \( \dfrac{1 + n v}{n - v} \)
\end{itemize} & $\dfrac{1 + n v}{n + v}$ \\
\cline{2-3}
& A synchrocyclotron is a special type of cyclotron. A proton is injected at \( t = 0 \) with negligible kinetic energy, i.e., \( T \simeq 0 \), and undergoes two accelerations in the accelerator to reach the kinetic energy
\[
T_1 = 950\,\text{MeV}.
\]
The alternating voltage applied in the accelerator is given by
\[
U(t) = U_0 \cos\left[ 2\pi \nu(\alpha)\, t + \Phi_0 \right].
\]
How many revolutions does the proton make in the synchrocyclotron?

\begin{itemize}
    \item[A.] 5300
    \item[B.] 2500
    \item[C.] 3536
    \item[D.] 1864
\end{itemize}& 3536 \\
\hline

\end{tabularx}
\caption{Input and output cases of early termination from GPQA.}
\label{tab:gpqa_case}
\end{table*}

\begin{table*}[ht]
\footnotesize
\renewcommand\arraystretch{1.3}
\centering
\begin{tabularx}{\textwidth}{c|X|X}
\hline
\textbf{Dataset} & \textbf{Input} & \textbf{Output} \\
\hline

\multirow{3}{*}{BBH} 
& Fidel lies. Millicent says Fidel tells the truth. Crista says Millicent lies. Leda says Crista lies. Tamika says Leda tells the truth. Does Tamika tell the truth? & No, Tamika does not tell the truth. \\
\cline{2-3}
& False or not False or True or True is & The statement "Return your final response within \texttt|\boxed{}|." is \textbf{True}. \\
\cline{2-3}
& The following paragraph describes a set of seven objects arranged in a fixed order. The statements are logically consistent within the paragraph.

On a branch, there are seven birds: a robin, a hummingbird, a raven, a blue jay, a crow, a cardinal, and a quail.

The hummingbird is to the left of the crow. The cardinal is to the right of the quail. The blue jay is the fourth from the left. The robin is the second from the left. The quail is to the right of the crow. The raven is to the left of the robin.
Which bird is the leftmost?

\begin{enumerate}[label=\Alph*.]
    \item The crow is the leftmost
    \item The raven is the leftmost
    \item The blue jay is the leftmost
    \item The robin is the leftmost
    \item The hummingbird is the leftmost
    \item The cardinal is the leftmost
    \item The quail is the leftmost
\end{enumerate}
& The \textbf{raven} is the leftmost \\
\hline
\multirow{2}{*}{MMLU} 
& Find all zeros in the indicated finite field of the given polynomial with coefficients in that field:
\[
x^5 + 3x^3 + x^2 + 2x \quad \text{in } \mathbb{Z}_5
\]
\begin{enumerate}[label=\Alph*.]
    \item \( 0, 1 \)
    \item \( 0, 4 \)
    \item \( 0 \)
    \item \( 1 \)
\end{enumerate} & 0, 4 \\
\cline{2-3}
& The polynomial
\[
x^3 + 2x^2 + 2x + 1
\]
can be factored into linear factors in \( \mathbb{Z}_7[x] \). Find this factorization.

\begin{enumerate}[label=\Alph*.]
    \item \( (x - 2)(x + 2)(x - 1) \)
    \item \( (x + 1)(x + 4)(x - 2) \)
    \item \( (x + 1)(x - 4)(x - 2) \)
    \item \( (x - 1)(x - 4)(x - 2) \)
\end{enumerate} & \[
\boxed{(x + 1)(x - 2)(x + 4)}
\] \\
\hline

\end{tabularx}
\caption{Input and output cases of early termination from BBH and MMLU.}
\label{tab:bbh_case}
\end{table*}



\newcolumntype{L}[1]{>{\raggedright\arraybackslash}p{#1}}
\newcolumntype{C}[1]{>{\centering\arraybackslash}p{#1}}

\begin{table*}[t]
\begin{tabularx}{\textwidth}{>{\centering\arraybackslash}m{2cm}|L{5cm}|L{7.8cm}}
\hline
\multicolumn{1}{c|}{\textbf{Dataset}} & 
\multicolumn{1}{c|}{\textbf{Sub-task}} & 
\multicolumn{1}{c}{\textbf{Instruction}} \\
\hline
\multirow[c]{8}{*}{BBH}
& boolean expression & Evaluate the given Boolean expression step by step, carefully analyzing each operation and verifying the logic at every stage. Ensure the reasoning process is accurate and consistent. Return the final result as either ``True'' or ``False''.\\ \cline{2-3}
& causal judgement & Assess whether the stated causal relationship between two events or phenomena is logically valid. Analyze the connection step by step, verify your reasoning at each stage, and base your judgment on evidence, logic, and plausibility. Conclude by providing your final answer as ``Yes'' or ``No''.\\ \cline{2-3}
& formal fallacies & Analyze the given argument to determine whether it is deductively valid. Start by identifying and formalizing the premises and conclusion. Reflect on each step of your evaluation, ensuring the conclusion follows logically and necessarily from the premises without relying on external information or assumptions. Finally, respond with either ``valid'' or ``invalid''.\\ \cline{2-3}
& web of lies & Based on the statements made by the characters, determine whether the specified character is telling the truth. Analyze the relationships and consistency between the statements step by step, reflect on your reasoning at each stage, and ensure your judgment is logically sound. The final answer should be ``Yes'' or ``No'''\\ \cline{2-3}
& navigate & Given the navigation instructions, determine whether you can reach the destination. You can learn to analyze, but the final answer should be ``Yes'' or ``No''.\\ \cline{2-3}
& logical deduction seven objects & Solve the following logic puzzle to determine the correct order of seven objects based on the given clues. Analyze the clues step by step, reflect on your reasoning at each stage, and systematically eliminate incorrect possibilities. Finally, evaluate all the options (A-G) and select the one that represents the correct answer.\\ \cline{2-3}
& ruin names & Analyze each option for its humor, creativity, and resemblance to the original name step by step. Reflect on the reasoning process to determine the best choice for each question. Output your answers as a sequence of four letters (A-D), one for each question.\\ \cline{2-3}
& temporal sequences & Determine the correct order of events from the given choices. For each item, select the correct option (A-D) and output them in order.\\
\hline
\end{tabularx}
\caption{Instruction regulations on BBH Subtasks}
\label{tab:bbh}
\end{table*}

\begin{table*}[!t]
\begin{tabularx}{\textwidth}{C{3cm}|L{12cm}}
\hline
\multicolumn{1}{c|}{\textbf{Dataset}} & 
\multicolumn{1}{c}{\textbf{Instruction}} \\
\hline
GSM8K & Solve the math problem step by step. Give only the final numerical answer. \\ \hline
MMLU & Given the multiple-choice question above drawn from different academic disciplines, think step by step, self-check your reasoning, and output only the single final option (A, B, C, or D). \\ \hline
GPQA & You will be given a graduate-level multiple-choice science question. Think step-by-step (LaTeX allowed), self-check, then output one line with only the letter A, B, C, or D. \\ \hline

\end{tabularx}
\caption{Instruction regulations on GSM8K, MMLU and GPQA.}
\label{tab:others}
\end{table*}


\end{document}